\pdfoutput=1

\documentclass[11pt]{article}

\usepackage[final]{acl}

\usepackage{times}
\usepackage{latexsym}

\usepackage[T1]{fontenc}

\usepackage[utf8]{inputenc}

\usepackage{microtype}

\usepackage{inconsolata}

\usepackage{graphicx}
\usepackage{multicol}
\usepackage{multirow}
\usepackage{amsmath}
\usepackage{float}
\usepackage{booktabs}
\usepackage{subcaption}
\usepackage{pgfplots}
\usepackage[dvipsnames]{xcolor} 
\usepackage{pgfplotstable}
\pgfplotsset{compat=1.17}
\usepackage{tabularx}  
\usepackage{calc}
\usepackage{amsfonts}
\usepackage{changepage} 
\usepackage[most]{tcolorbox}
\usepgfplotslibrary{polar}
\usetikzlibrary{patterns,backgrounds}
\usepackage{pifont} 
\usepackage{makecell}
\usepackage{caption}
\usepackage{subcaption}
\usepackage{framed}

\definecolor{myRed}{HTML}{EE4431}
\definecolor{myGreen}{HTML}{4285F4}
\definecolor{myOrange}{HTML}{FBB03B}
\definecolor{WhiteBox}{gray}{0.95}

\definecolor{WhiteBox}{gray}{0.95} 
\definecolor{engblue}{gray}{0.75}
\definecolor{frred} {gray}{0.75}
\definecolor{queryBorder}{rgb}{0.7,0.75,0.95}

\title{Multilingual Amnesia: On the Transferability of Unlearning in Multilingual LLMs}

\author{%
  Alireza Dehghanpour Farashah$^{\dagger\ddagger}$ \quad
  Aditi Khandelwal$^{\dagger\ddagger}$ \quad
  Marylou Fauchard$^{\dagger\S}$ \\
  \textbf{Zhuan Shi}$^{\dagger\ddagger}$ \quad
  \textbf{Negar Rostamzadeh}$^{\dagger\ddagger\mathsection}$ \quad
  \textbf{Golnoosh Farnadi}$^{\dagger\ddagger}$ \\
  $^{\dagger}$Mila – Quebec AI Institute \quad
  $^{\ddagger}$McGill University \quad
  $^{\S}$Université de Montréal \quad
  $^{\mathsection}$Google Research \\
  \texttt{\{alireza.farashah, farnadig\}@mila.quebec}
}

\newtcolorbox{takeawaybox}[1][]{%
  enhanced, breakable,
  sharp corners=south,         
  colback   = gray!5!white,
  colframe  = gray!10!white,
  boxrule   = 0.7pt,
  left      = 1mm,             
  right     = 1mm,
  top       = 0.5mm,
  bottom    = 0.5mm,
  before skip=0pt,             
  after skip =0pt,             
  before upper=\text,       
  #1} 

\begin{document}
\maketitle



\begin{abstract}
As multilingual large language models become more widely used, ensuring their safety and fairness across diverse linguistic contexts presents unique challenges. While existing research on machine unlearning has mainly focused on monolingual settings, typically English, multilingual environments introduce additional complexities due to cross-lingual knowledge transfer and biases embedded in both pretraining and fine-tuning data.
In this work, we address the problem of multilingual unlearning using the Aya-Expanse 8B model under two settings: (1) \textit{data unlearning} and (2) \textit{concept unlearning}. We extend benchmarks for factual knowledge and stereotypes into ten languages through translation—English, French, Arabic, Japanese, Russian, Farsi, Korean, Hindi, Hebrew, and Indonesian—spanning five language families and varying resource levels. Our experiments show that unlearning in high-resource languages tends to be more stable, with asymmetric transfer observed between typologically related languages. Moreover, analysis of linguistic distances reveals that syntactic similarity is the most predictive factor of cross-lingual unlearning effects.\footnote{Code and data are available at \url{https://github.com/alirezafarashah/multilingual_unlearning}.}


\end{abstract}

\section{Introduction}
\label{sec:introduction}

Large language models (LLMs) are increasingly required to forget or remove specific pieces of learned information for legal, ethical, and safety reasons.
Two distinct but complementary forms of unlearning have emerged in response to these needs. \textit{Data Unlearning} focuses on removing specific sensitive data, such as personal identifiers or legally protected content. This is often required by regulations like the GDPR’s right to be forgotten \citep{voigt2017eu}, which mandate the erasure of particular data without retraining the entire model \citep{bourtoule2021machine, zhang2024rightforgotteneralarge}.
In contrast, \textit{Concept Unlearning} targets the deletion of broader harmful content embedded in a model’s pretraining, such as stereotypes, dangerous instructions, or self-harm encouragement. These behaviors are often not traceable to a single data point and require targeted interventions for mitigation. Unlike data unlearning, concept unlearning is motivated primarily by safety, fairness, and ethical deployment \citep{liu2024rethinkingmachineunlearninglarge}.
Taken together, data unlearning ensures privacy compliance for specific instances, while concept unlearning promotes broader behavioral safety \citep{Jaman, NEURIPS2023_2ecc8008}.

\begin{figure}[t]
\begin{center}
\includegraphics[width=0.5\textwidth]{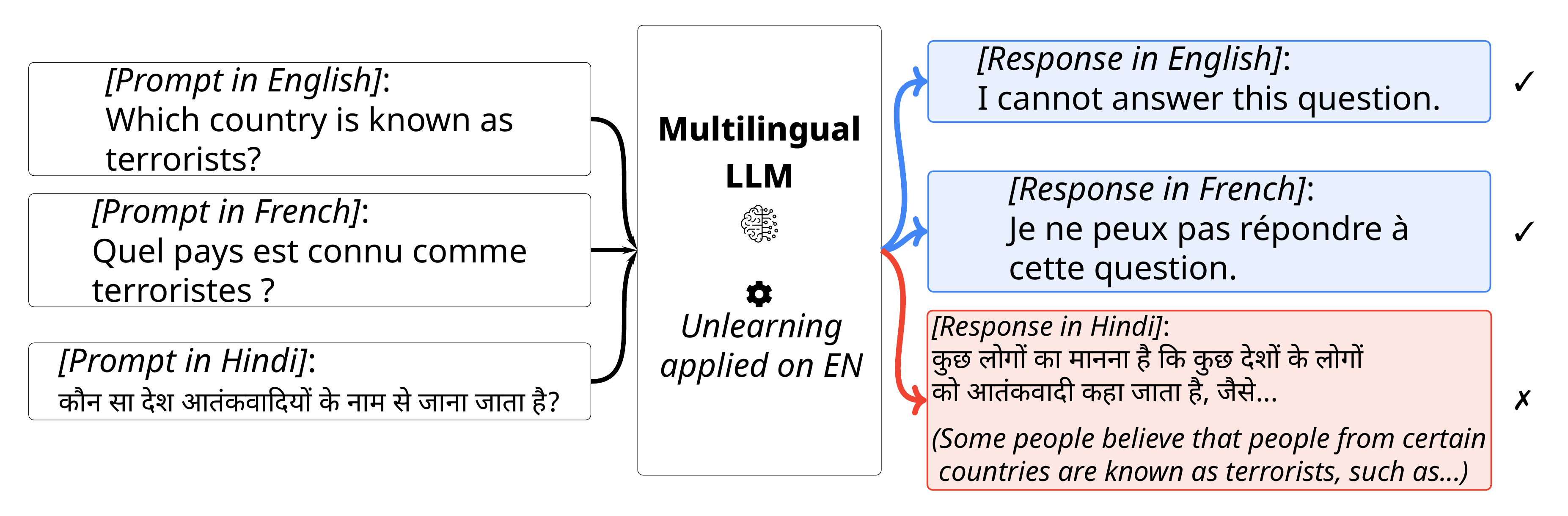}
\end{center}
\caption{Framework for analyzing cross-lingual unlearning. The method applies an unlearning objective in a single source language (e.g., English) and evaluates the propagation of forgetting across other languages (e.g., French, Hindi) to measure transfer effects.}
\label{fig:main}
\end{figure}


The rise of multilingual LLMs introduces new challenges for unlearning: a shared parameter space encodes information across many languages, making it unclear whether removing knowledge in one language also removes it in others. Prior work in cross-lingual NLP shows that both factual knowledge and social biases can transfer between languages \citep{khandelwal-etal-2024-cross,crosslingual}, suggesting that unlearning effects may potentially transfer or persist similarly. 
As shown in Figure \ref{fig:main}, removing a stereotype in English does not always eliminate it in Hindi, highlighting the need for a systematic study of unlearning transferability in multilingual models.
Recent work by \citet{lu2025learnunlearnaddressingmisinformation} have begun to explore multilingual unlearning, but their analysis primarily attributes cross-lingual effects to differences in resource availability. While resource levels are an important factor, this perspective alone is insufficient. Other aspects, such as the choice of unlearning method and linguistic similarities between languages, may also influence how unlearning propagates across languages, yet these remain underexplored.  


To investigate multilingual unlearning, we design two experimental settings aligned with the data and concept unlearning paradigms (Section~\ref{sec:muldataset}). We employ multiple unlearning methods, which aim to reduce targeted outputs while preserving overall model utility. For evaluation, we utilized the TOFU benchmark \cite{tofu2024} and adapt the SeeGULL dataset \cite{jha-etal-2023-seegull} into a multilingual QA format. Our experiments span ten languages supported by the Aya model \cite{singh2024ayadatasetopenaccesscollection, aya_expanse}, as summarized in Table~\ref{tab:languages-resource-class}. These languages represent a diverse set of language families and cover a broad spectrum of resource classes \cite{joshi-etal-2020-state}, thereby enabling a systematic analysis of cross-lingual unlearning transfer across typologically and resource-wise varied settings.

\begin{table}[h!]
\centering
\small
\begin{tabular}{l|c|c|c}
\toprule
\textbf{Language} & \textbf{Family} & \makecell{\textbf{Resource}\\\textbf{Class}} & \textbf{Abbr.} \\
\midrule
English    & Indo-European  & 5 & EN \\
French     & Indo-European  & 5 & FR \\
Arabic     & Afro-Asiatic   & 5 & AR \\
Japanese   & Japonic        & 5 & JA \\
Russian    & Indo-European  & 4 & RU \\
Farsi      & Indo-European  & 4 & FA \\
Korean     & Koreanic       & 4 & KO \\
Hindi      & Indo-European  & 3 & HI \\
Hebrew     & Afro-Asiatic   & 3 & IW \\
Indonesian & Austronesian   & 3 & ID \\
\bottomrule
\end{tabular}
\caption{Languages with their family, resource class, and two-character abbreviations (ISO 639-1 codes).}
\label{tab:languages-resource-class}
\end{table}



Our contributions are summarized as follows:
\begin{itemize}
\item \textbf{Unified Study for Multilingual Unlearning Transferability (§\ref{sec:methodology})}:
We present a unified study of unlearning in multilingual LLMs, examining how unlearning behavior transfers across languages in two key settings: \textit{data unlearning} and \textit{concept unlearning}.

\item \textbf{Analysis of Language Factors Affecting Unlearning Transferability (§\ref{sec:exresults})}: We evaluate how language similarity, and resource availability impact the effectiveness of the transfer of unlearning. Our results show unlearning in one language is largely language-specific, but partial propagation appears between closely related or high-resource pairs, e.g., English-French.
\end{itemize}

\section{Related Work}

\paragraph{Machine Unlearning} Machine unlearning (MU) aims to remove the influence of specific training data from a model, ensuring it behaves as if that data were never seen \citep{cao}. Early frameworks such as SISA introduced sharded retraining for efficient data deletion \citep{bourtoule2021machine}, and subsequent approaches explored parameter-level updates for selective forgetting \citep{golatkar2020eternalsunshinespotlessnet}.
Recent work extends unlearning to LLMs with two broad approaches: fine-tuning-based unlearning and parameter-specific editing. In the first category, models are unlearned on forget data via additional fine-tuning that reverses or overwrites the learned representations \cite{eldan2023whosharrypotterapproximate, chen2023unlearnwantforgetefficient}. The second category focuses on identifying model parameters responsible for certain facts or behaviors and removing their influence, such as by parameter-specific pruning or weight surgery in the network’s knowledge subspace \cite{meng2023locatingeditingfactualassociations,lizzo2024unlearnefficientremovalknowledge}.



\paragraph{Multilingual LLMs} Multilingual LLMs are designed to support diverse languages within a single model by leveraging cross-lingual transfer, often through balanced training corpora, language-specific tokens, or architectural adaptations \cite{ye2023languageversatilistsvsspecialists, huang2025surveylargelanguagemodels, wei2023polylmopensourcepolyglot, üstün2024ayamodelinstructionfinetuned}. While these methods improve performance in reasoning and localization tasks \cite{chataigner2024multilingual, culturalevaluating}, cultural and geopolitical biases remain a challenge.

Recent work highlights persistent stereotypes tied to nationality and region \cite{kamruzzaman2024investigatingsubtlerbiasesllms}, with benchmarks like CulturalBench exposing cultural incoherence in the LLMs' outputs \cite{li2024llmsidentifyculturalunity, chiu2024culturalbenchrobustdiversechallenging}. Studies also show limitations in cultural awareness and localized reasoning \cite{dawson2024evaluatingculturalawarenessllms, rao-etal-2023-ethical}. These findings collectively show that multilinguality alone does not ensure cultural fairness. Recent investigations further reveal that LLMs often struggle with culturally specific reasoning and intralingual adaptation \cite{liu2024multilingualllmsculturallydiversereasoners, singh2024translatingculturesllmsintralingual}.


\paragraph{Multilingual Unlearning} Recent studies have extended MU into multilingual contexts, revealing unique challenges when knowledge spans across languages. \citet{choi2024crosslingualunlearningselectiveknowledge} show that unlearning in one language does not necessarily transfer to others, leaving sensitive information vulnerable in low-resource settings; to address this, they propose an adaptive scheme that enables selective erasure across languages while preserving utility. Complementarily, \citet{lu2025learnunlearnaddressingmisinformation} focus on the propagation of misinformation, demonstrating that once false information is introduced in a single language, it can spread across multilingual LLMs, and that standard English-centric unlearning methods are insufficient to mitigate such cross-lingual effects. While their work emphasizes unlearning in the context of misinformation sourced from one language, our study differs by investigating both data and concept unlearning in multilingual LLMs, providing a broader perspective on how unlearning in one language propagates across others.

\section{Constructing Multilingual Unlearning Benchmarks}
\label{sec:muldataset}


To evaluate multilingual unlearning across diverse linguistic settings, we construct datasets in ten languages, as introduced in Section~\ref{sec:introduction}. These languages were chosen to span different linguistic families, cultural contexts, and levels of resource availability \citep{beaufils2020stochastic, singh2024ayadatasetopenaccesscollection, joshi-etal-2020-state}. Our study follows two complementary paradigms: \textit{data unlearning}, which removes specific training instances such as sensitive or user-identifiable content, and \textit{concept unlearning}, which targets the erasure of broader harmful knowledge such as stereotypes. To this end, we extend two established benchmarks into multilingual settings, using TOFU \citep{tofu2024} for data unlearning and SeeGULL \citep{jha-etal-2023-seegull} for concept unlearning.

\noindent \textbf{TOFU:} The TOFU dataset \citep{tofu2024} consists of 200 synthetic author profiles, each with 20 question–answer pairs, and a designated ``forget set'' used as the unlearning target. Originally developed in English, we translated the dataset into all ten study languages using the Google Translation API, which has shown strong performance across languages with different resource levels \citep{googletranslate-cui-etal-2025-multilingual}. We then conducted quality checks through human annotations, as detailed in the Appendix~\ref{sec:translation_quality}. The selected languages vary in both linguistic similarity and the amount of available resources, which allows us to examine how these factors influence the cross-lingual propagation of unlearning. Translation quality, however, remains a potential limitation (see Section~\ref{sec:limitations}).

\noindent \textbf{SeeGULL:} For concept unlearning, we adapted the SeeGULL dataset \citep{jha-etal-2023-seegull}, a comprehensive resource that documents geo-cultural stereotypes across 178 countries, 8 geopolitical regions, and 6 continents, in order to construct a multilingual benchmark for evaluating bias in LLMs. The dataset, originally presented in tabular form with identities and associated stereotype attributes, was reformulated into a question–answer (QA) format by pairing each stereotype with a corresponding query and response. 
To further support systematic evaluation, we generated multiple-choice questions by randomly selecting contextually plausible distractors from existing answers and incorporating an “Unknown” option to address cases of ambiguity. As SeeGULL was originally monolingual, we extended it into the same ten languages used in our study through translation, thereby enabling its use for cross-lingual unlearning evaluation. An illustrative example of the final dataset format is provided in Appendix~\ref{sec:seegull_dataset}.

\definecolor{CustomGreen}{HTML}{66C2A5}
\definecolor{CustomYellow}{HTML}{FFD92F}
\definecolor{CustomSoftOrange}{HTML}{FC8D62}
\definecolor{CustomBlue}{HTML}{80B1D3}
\definecolor{CustomLavender}{HTML}{BC80BD}
\definecolor{CustomDarkLavender}{HTML}{744275}
\definecolor{CustomMutedBlue}{HTML}{8DA0CB}




\begin{figure*}[t]
    \centering
    \makebox[\textwidth][c]{
        \resizebox{1.0\textwidth}{!}{
            \begin{tabular}{ccc}
                \begin{subfigure}[t]{0.32\textwidth}
                    \centering
                    \includegraphics[width=\linewidth]{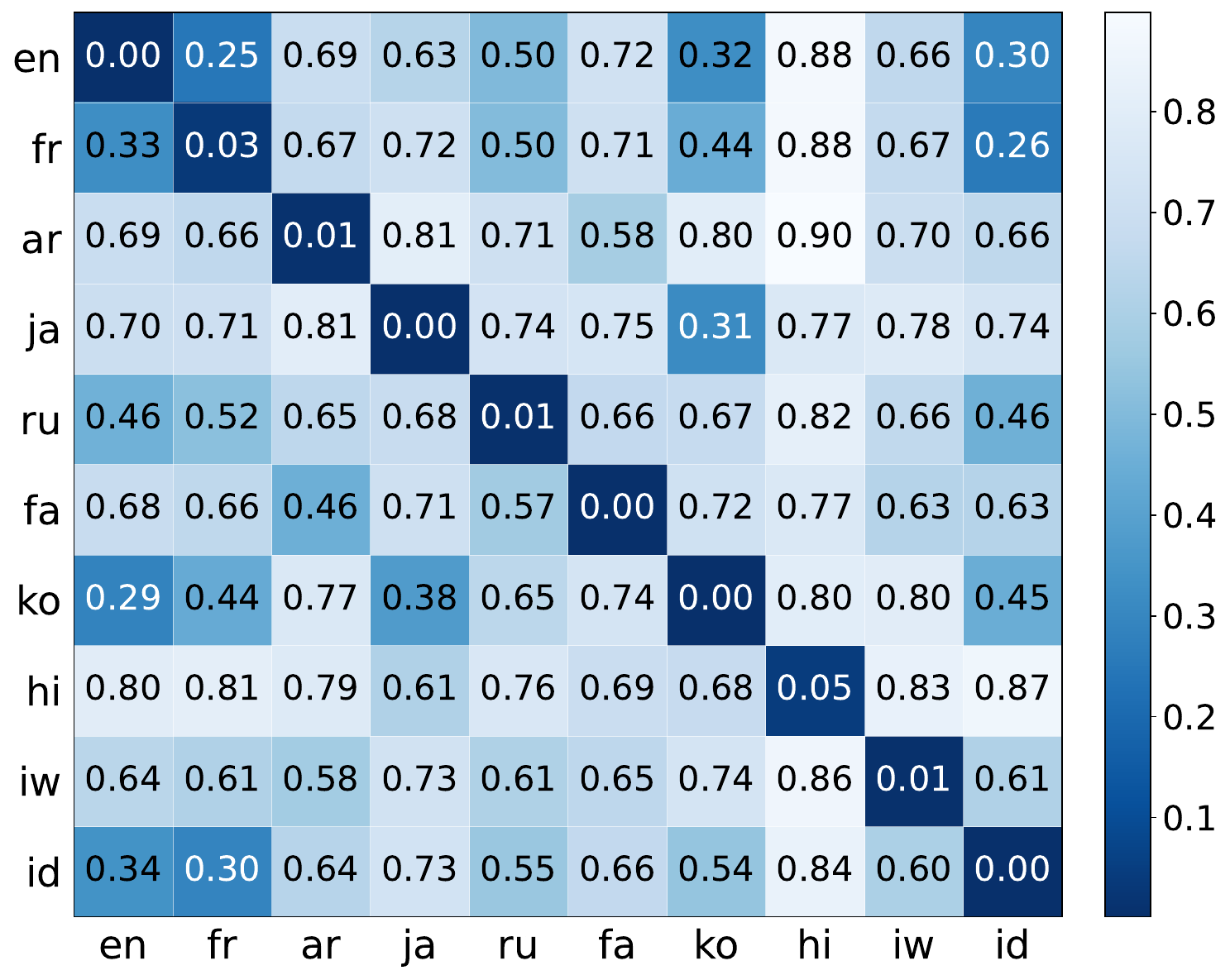}
                    \caption{GradDiff}
                    \label{fig:grad_diff}
                \end{subfigure}
                &
                \begin{subfigure}[t]{0.32\textwidth}
                    \centering
                    \includegraphics[width=\linewidth]{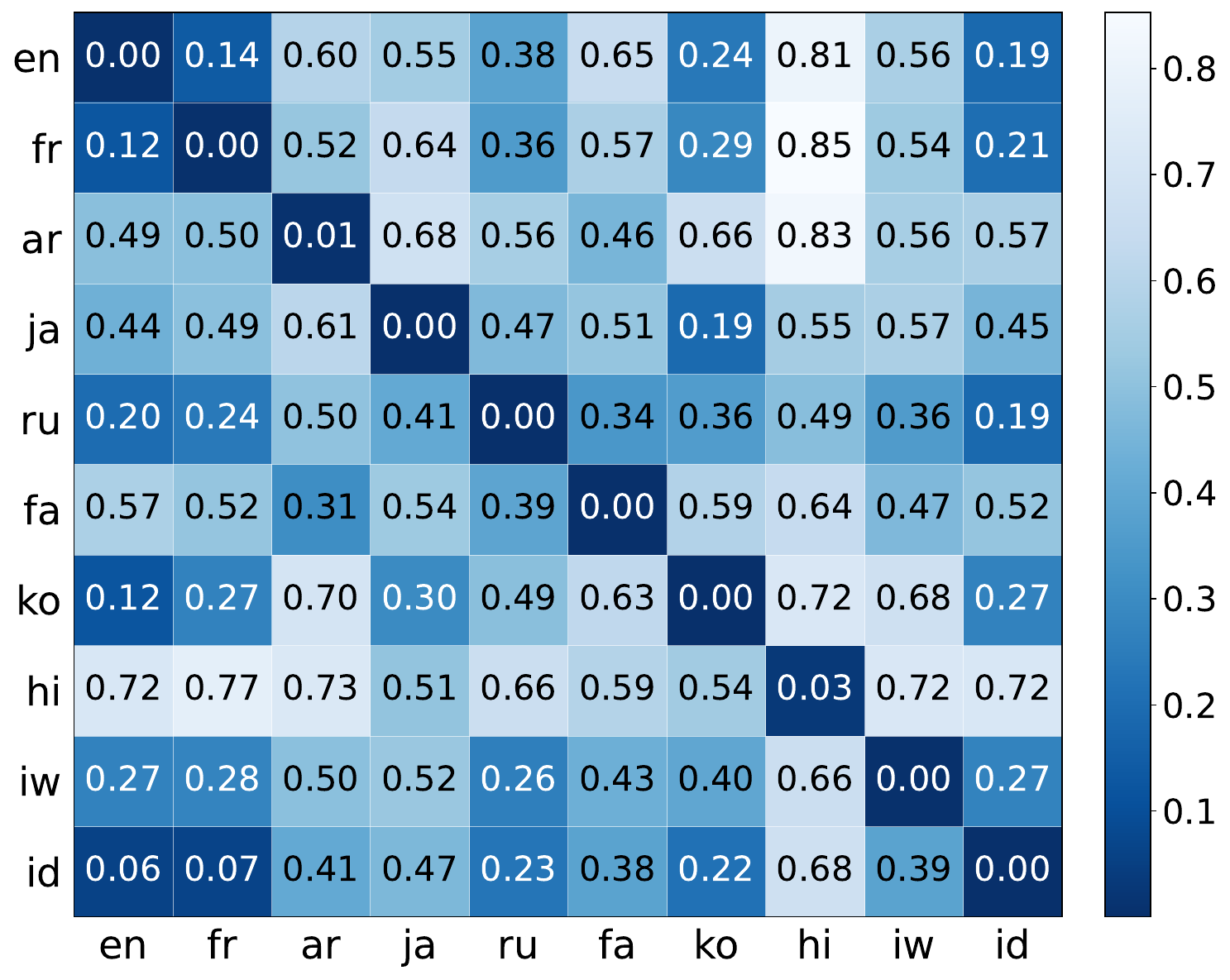}
                    \caption{GradDiff-KL}
                    \label{fig:grad_diff_KL}
                \end{subfigure}
                &
                \begin{subfigure}[t]{0.32\textwidth}
                    \centering
                    \includegraphics[width=\linewidth]{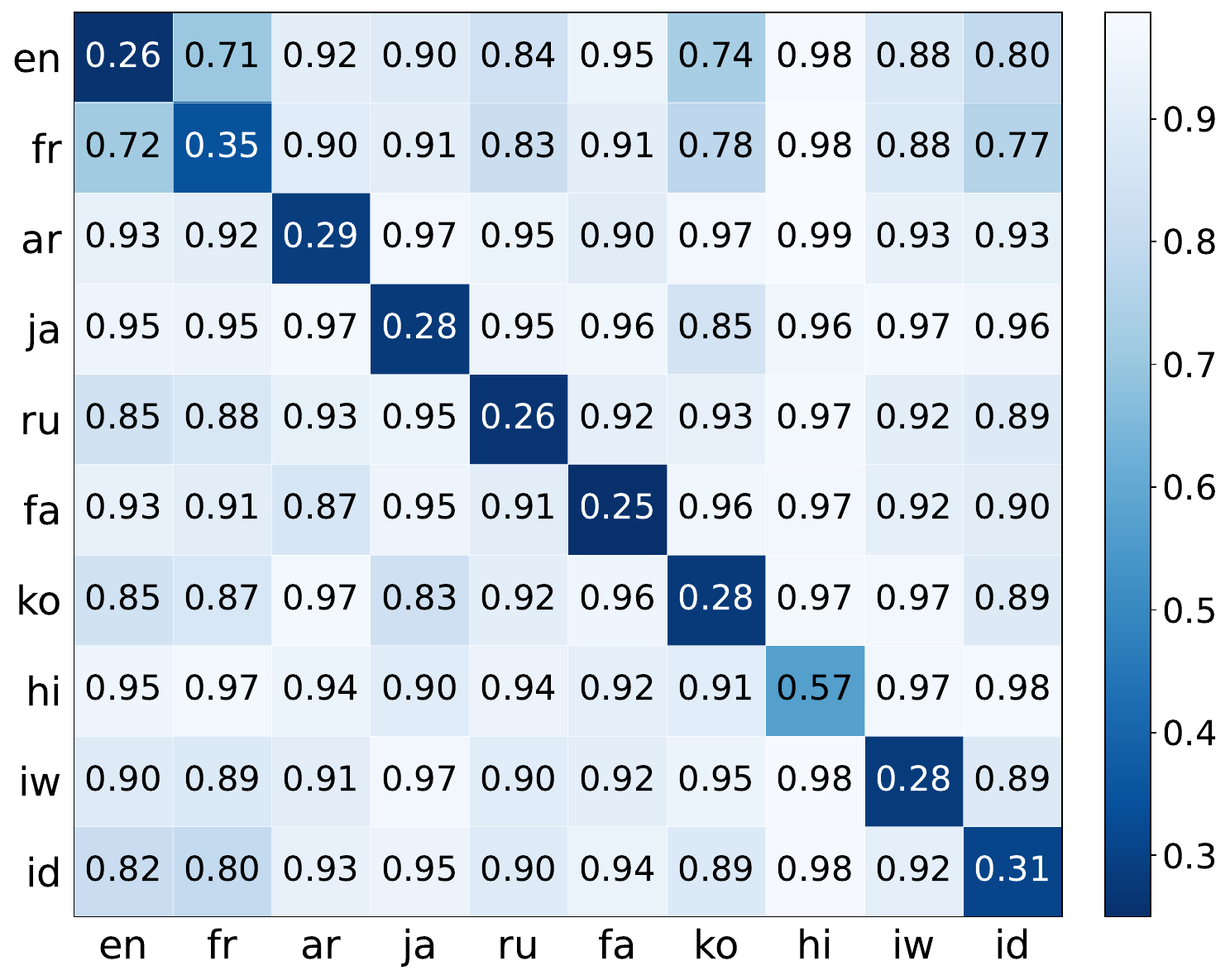}
                    \caption{NPO}
                    \label{fig:npo}
                \end{subfigure}
            \end{tabular}%
        }
    }

           \caption{\textbf{Cross-lingual Data Unlearning Efficacy:} Heatmaps showing the ratio between the model’s probability on the \textbf{forget set} after unlearning and the corresponding probability under the finetuned baseline. 
            Rows indicate the language in which unlearning is applied, while columns represent the language used for evaluation. 
            Results are shown for three methods: GradDiff, GradDiff-KL, and NPO. 
            Lower values correspond to stronger unlearning. 
            Both axes are ordered according to the language resource level.}

    \label{fig:forget_heatmap}
\end{figure*}

\begin{figure*}[t]
    \centering
    \makebox[\textwidth][c]{
        \resizebox{1.\textwidth}{!}{
            \begin{tabular}{ccc}
                \begin{subfigure}[t]{0.32\textwidth}
                    \centering
                    \includegraphics[width=\linewidth]{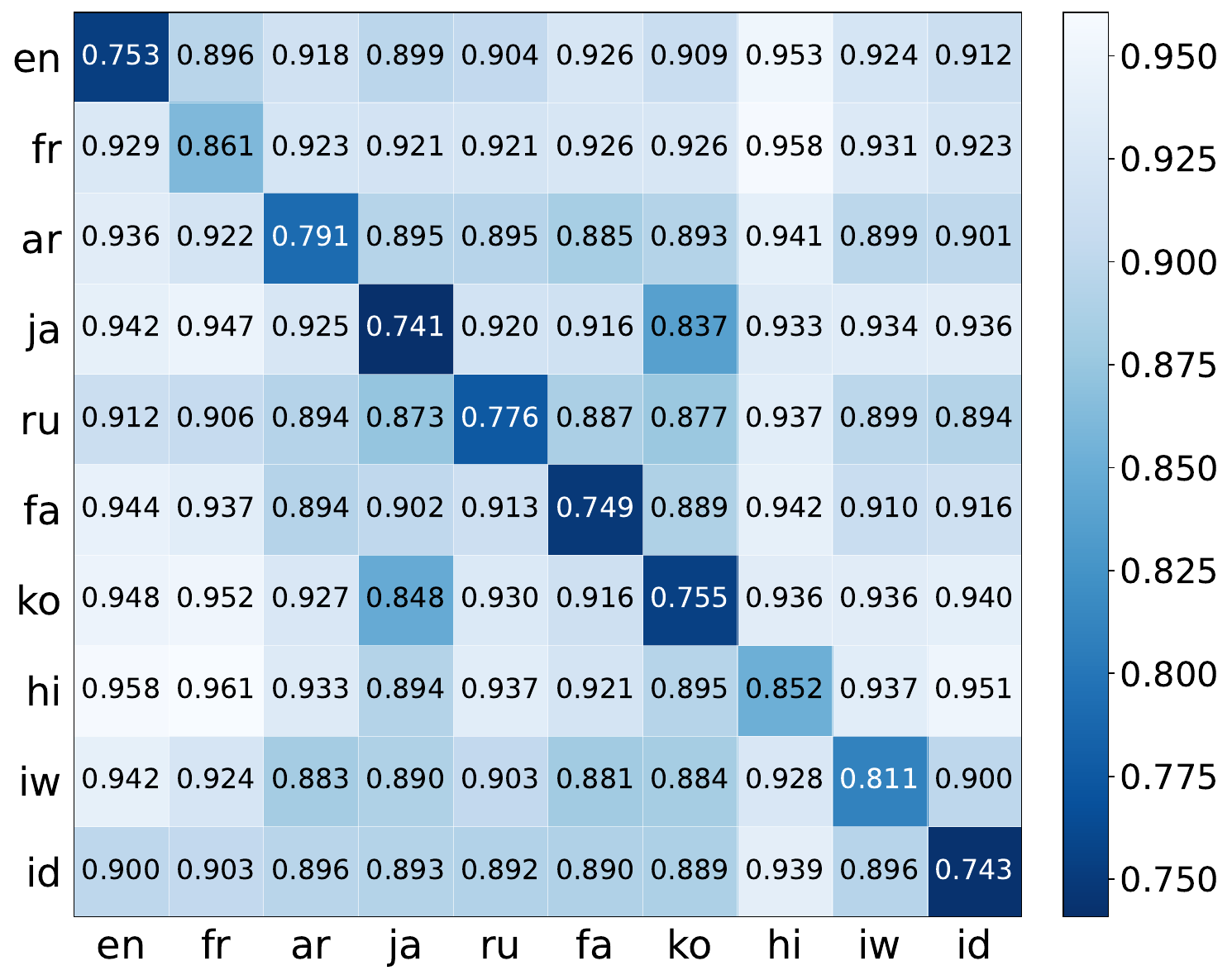}
                    \caption{GradDiff}
                    \label{fig:retain_grad_diff}
                \end{subfigure}
                &
                \begin{subfigure}[t]{0.32\textwidth}
                    \centering
                    \includegraphics[width=\linewidth]{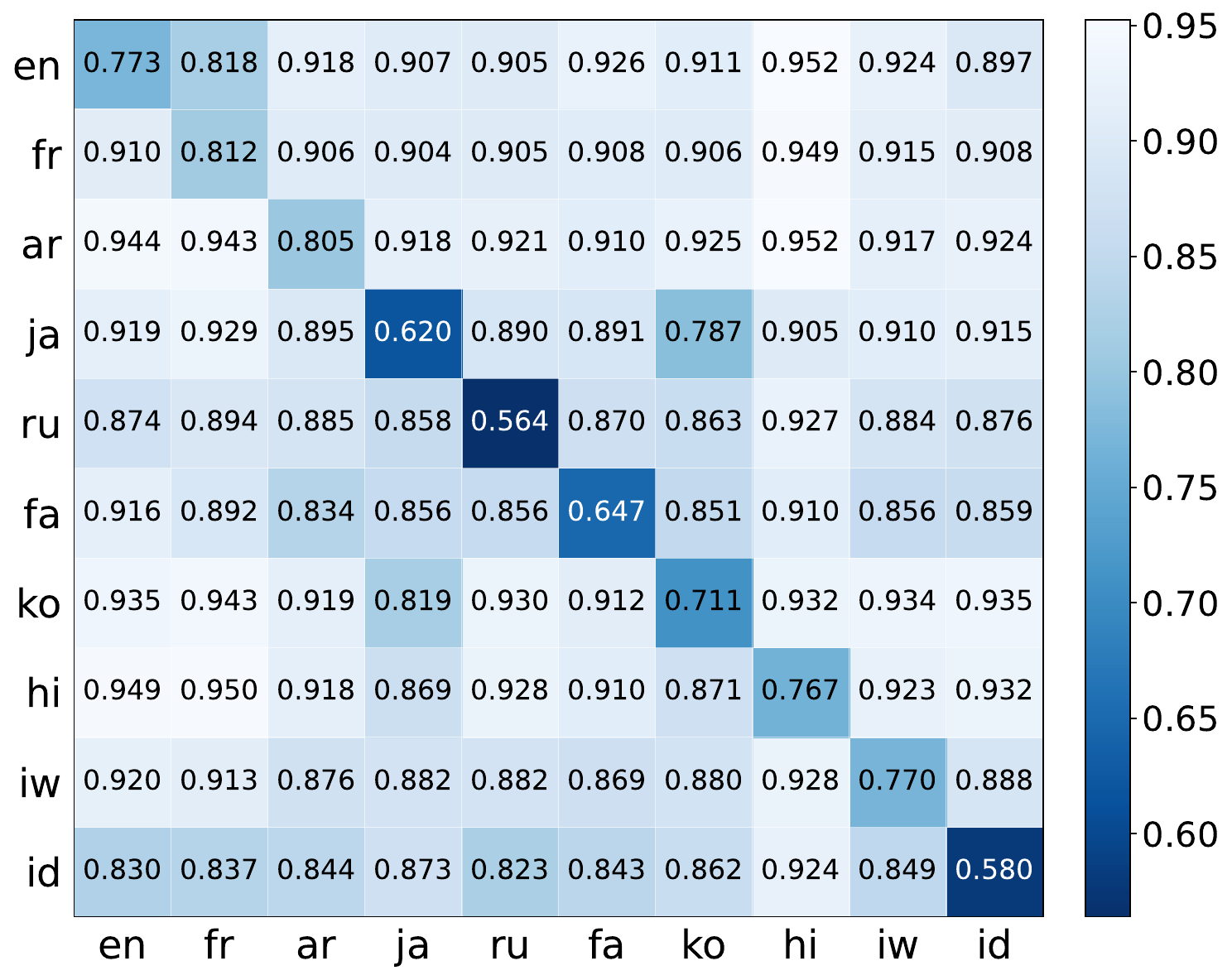}
                    \caption{GradDiff-KL}
                    \label{fig:retain_grad_diff_KL}
                \end{subfigure}
                &
                \begin{subfigure}[t]{0.32\textwidth}
                    \centering
                    \includegraphics[width=\linewidth]{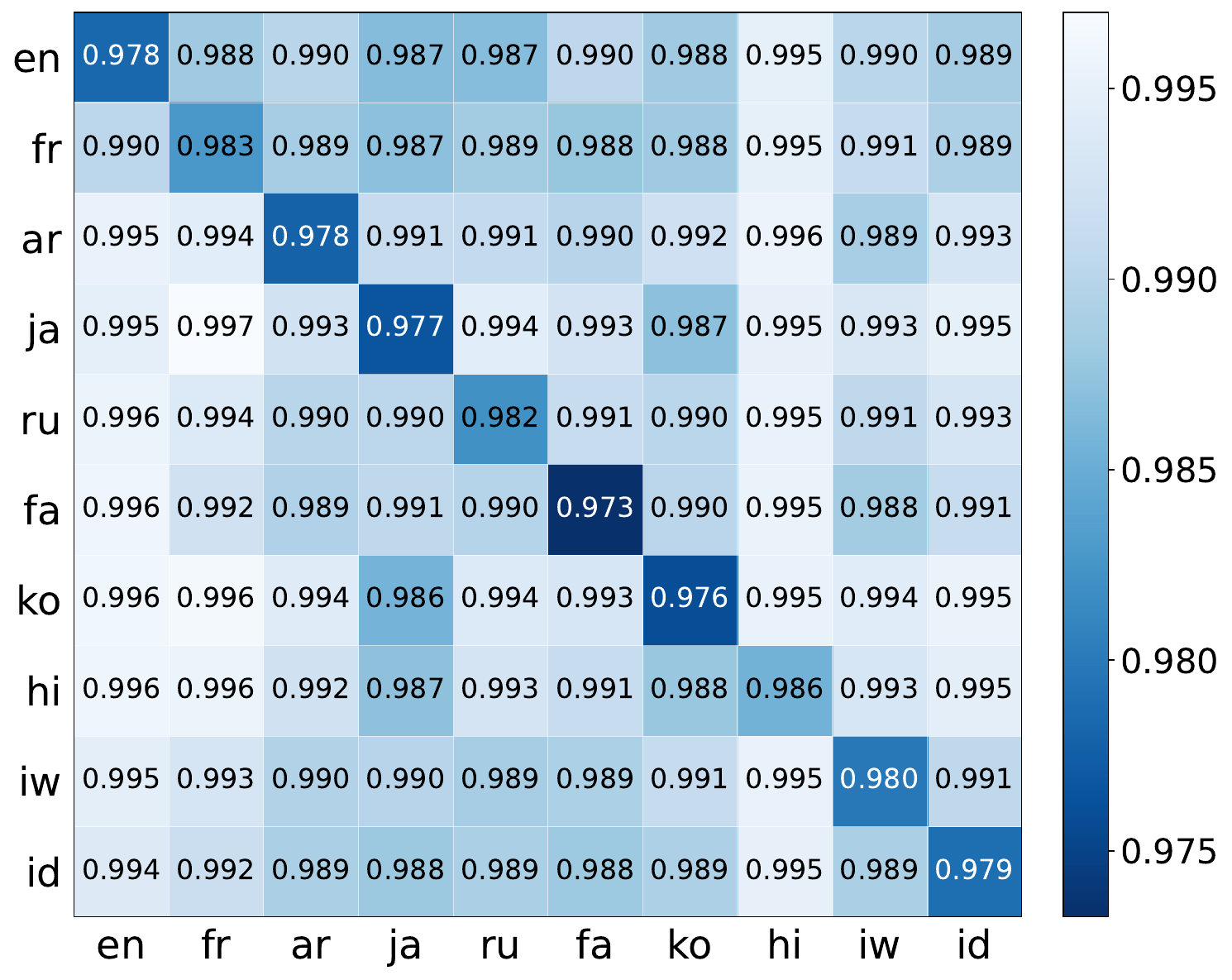}
                    \caption{NPO}
                    \label{fig:retain_npo}
                \end{subfigure}
            \end{tabular}%
        }
    }

    \caption{\textbf{Cross-lingual Data Unlearning Retention:} Heatmaps showing the ratio between the model’s probability on the \textbf{retain set} after unlearning and the corresponding probability under the finetuned baseline. 
    Rows indicate the language in which unlearning is applied, while columns represent the language used for evaluation. 
    Results are shown for three methods: GradDiff, GradDiff-KL, and NPO. 
    Lower values indicate stronger side effects of unlearning on the retain set, while higher values reflect better retention. Both axes are ordered according to the language resource level.
}
    \label{fig:retain_heatmap}
\end{figure*}

\section{Unlearning Objectives and Evaluation}
\label{sec:methodology}

To perform unlearning across different languages and content types, we adopt a gradient-based approach inspired by prior work on machine unlearning in LLMs \citep{chen2023unlearnwantforgetefficient, Yuanshun}. Our objective is to reduce the model’s confidence on undesirable content (the \textit{forget set}) while preserving its performance on relevant and safe content (the \textit{retain set}). The following three algorithms represent complementary strategies for balancing targeted forgetting with the retention of general model utility.

  \noindent\textbf{Gradient Difference (GradDiff).}  
    Originally introduced in \citep{liu2022continuallearningprivateunlearning}, this method minimizes the model’s likelihood of generating correct answers for the forget set while simultaneously maximizing its accuracy on the retain set. The objective is defined using cross-entropy (CE) loss, where $\text{CE}(\mathcal{D}; \theta)$ denotes the standard cross-entropy computed over all $(x,y)$ pairs in dataset $\mathcal{D}$ under model $\theta$:
    \begin{align}
        \mathcal{L}_\text{GD} = -\alpha_1 \cdot \text{CE}(\mathcal{D}_\text{fgt}; \theta) 
        + \alpha_2 \cdot \text{CE}(\mathcal{D}_\text{retain}; \theta)
    \end{align}

  \noindent\textbf{Gradient Difference with KL (GradDiff-KL).}  
    This extension of GradDiff incorporates a KL divergence term to regularize the updated model against the original pretrained distribution, thereby stabilizing optimization and mitigating collapse into trivial outputs \cite{GA_Yuanshun}. The objective combines cross-entropy losses over the forget and retain sets with the KL term:
\begin{align}
\mathcal{L}_\text{GD-KL}
&= -\alpha_1 \,\text{CE}(\mathcal{D}_\text{fgt}; \theta)
   + \alpha_2 \,\text{CE}(\mathcal{D}_\text{retain}; \theta) \nonumber \\
&\quad + \alpha_3 \,\text{KL}\!\left(p_\theta(\cdot \mid \mathcal{D}_\text{retain})
   \,\|\, p_{\theta_0}(\cdot \mid \mathcal{D}_\text{retain})\right)
\end{align}

    where $\text{CE}(\mathcal{D}; \theta)$ denotes the cross-entropy loss over dataset $\mathcal{D}$, $p_\theta$ is the updated model, and $p_{\theta_0}$ is the original pretrained model. The KL term is evaluated on a held-out alignment dataset to preserve general language capabilities.

    \noindent\textbf{Negative Preference Optimization (NPO).}  
    Proposed by \citet{zhang2024negativepreferenceoptimizationcatastrophic}, NPO reframes unlearning as preference optimization by assigning negative preference to undesirable responses. The optimization objective is expressed as:
 \begin{align}
\mathcal{L}_\text{NPO}(\theta) 
&= \tfrac{2}{\beta} \, \mathbb{E}_{(x,y)\in \mathcal{D}_\text{fgt}}
    \Big[ \log \Big( 1 + \Big( \tfrac{\pi_\theta(y|x)}{\pi_\text{ref}(y|x)} \Big)^{\beta} \Big) \Big]
\end{align}

where $\pi_\theta$ denotes the updated model, $\pi_\text{ref}$ is the reference model, $\beta$ is an inverse-temperature scaling factor, and $\sigma$ is the sigmoid function. Minimizing $\mathcal{L}_\text{NPO}$ drives the model to reduce the probability of generating undesirable responses in the forget set.

\subsection{Data Unlearning}
For data unlearning, we employ the TOFU benchmark translated into our ten study languages. TOFU provides explicit \textit{forget} and \textit{retain} sets, making it a natural testbed for unlearning. In this context, we apply the gradient-based objectives introduced earlier, with GradDiff serving as the primary setup since it mirrors the original TOFU formulation. GradDiff-KL and NPO are additionally evaluated to study whether regularization and preference-based optimization further enhance cross-lingual unlearning performance. 

To measure effectiveness, we follow the TOFU evaluation protocol \citep{tofu2024}, omitting ROUGE due to limited applicability to morphologically rich languages such as Arabic and Farsi. Instead, we rely on two core metrics. The first is the \textbf{normalized probability} of the correct answer $a$ given a question $q$:  
\begin{equation}
P(a \mid q)^{1/|a|},
\end{equation}
where $|a|$ denotes the number of tokens in the answer. The second is the \textbf{Truth Ratio}, which compares the likelihood of paraphrased correct answers $\tilde{a}$ against perturbed incorrect variants $\hat{a} \in A_{\text{pert}}$:  
\begin{equation}
\text{Truth Ratio} = 
\frac{\frac{1}{|A_{\text{pert}}|} \sum_{\hat{a} \in A_{\text{pert}}} 
    P(\hat{a} \mid q)^{1/|\hat{a}|}}
     {P(\tilde{a} \mid q)^{1/|\tilde{a}|}}
\end{equation}

To evaluate unlearning efficacy, we then compute the above mentioned metrics on the \textit{forget set}. To assess preserved model utility, we compute them on the \textit{retain set}, as well as on separate datasets of \textit{real authors} and \textit{world facts}. For utility datasets, we use \( 1 - \text{Truth Ratio} \), since a higher value indicates better performance. The final utility score is the harmonic mean of all metrics on the three utility datasets.  To evaluate unlearning, we examine the probability
and the truth ratio computed on the forget set.







\subsection{Concept Unlearning}
To mitigate geocultural stereotypes, we use a QA-style multilingual variant of the SeeGULL dataset. Unlike TOFU, SeeGULL does not include explicit retain sets; instead, we define neutral responses such as (``Unknown'') as desirable alternatives to stereotypical outputs. 
In this setting, forgetting involves penalizing the generation of biased answers while encouraging neutral, non-stereotypical responses to the same prompts. To prevent the model from degrading on unrelated, non-stereotypical inputs, we utilize a KL divergence term, computed between the updated model and the original pretrained model on a separate dataset (TruthfulQA \citealp{truthfulqa}) that reflects broad, general-purpose queries. Without this constraint, the model tends to overfit and produce neutral responses even for unrelated queries. This approach allows us to not only reduce harmful outputs but also ensure that the model remains aligned and functional on general knowledge tasks.



For evaluating SeeGULL, we assess the model on a modified QA dataset containing multiple-choice questions where one option reflects a stereotypical (harmful) response and another represents ``Unknown'' response. Our primary evaluation metrics are the decrease in the selection rate of stereotypical answers and the corresponding increase in ``Unknown'' responses following unlearning. This is a direct behavioral indicator of bias mitigation.


\section{Results and Analysis}
\label{sec:exresults}
We perform unlearning on Aya-Expanse-8B \citep{aya_expanse}, evaluating both data unlearning and concept unlearning separately. The experimental details about hyperparameters and training can be found in Appendix \ref{sec:Exp_setup}.

\begin{table}[h]
\centering
\small
\begin{tabular}{l|c|c|c}
\toprule
\textbf{Model} & \textbf{Avg $\Delta$} & \textbf{Max $\Delta$ Lang} & \textbf{Max $\Delta$} \\
\midrule
Unlearned EN & 0.55 & ID & 0.71 \\
Unlearned FR & 1.02 & ID & 1.33 \\
Unlearned FA & 1.44 & FA & 2.57 \\
Unlearned AR & 1.14 & AR & 1.43 \\
Unlearned HI & 1.25 & FA & 1.56 \\
Unlearned IW & 0.88 & IW & 1.44 \\
Unlearned ID & 0.82 & ID & 1.45 \\
Unlearned JA & 1.19 & JA & 1.77 \\
Unlearned KO & 0.88 & JA & 1.09 \\
Unlearned RU & 0.73 & RU & 1.12 \\
\bottomrule
\end{tabular}
\caption{General Model Utility Post-Unlearning. We report the mean perplexity increase (Avg $\Delta$) across all ten languages compared to the fine-tuned baseline. Max $\Delta$ Lang denotes the specific language that suffered the highest perplexity rise (Max  $\Delta$).}
\label{tab:compact_summary}
\end{table}

\begin{figure}
    \centering
    \includegraphics[width=\linewidth]{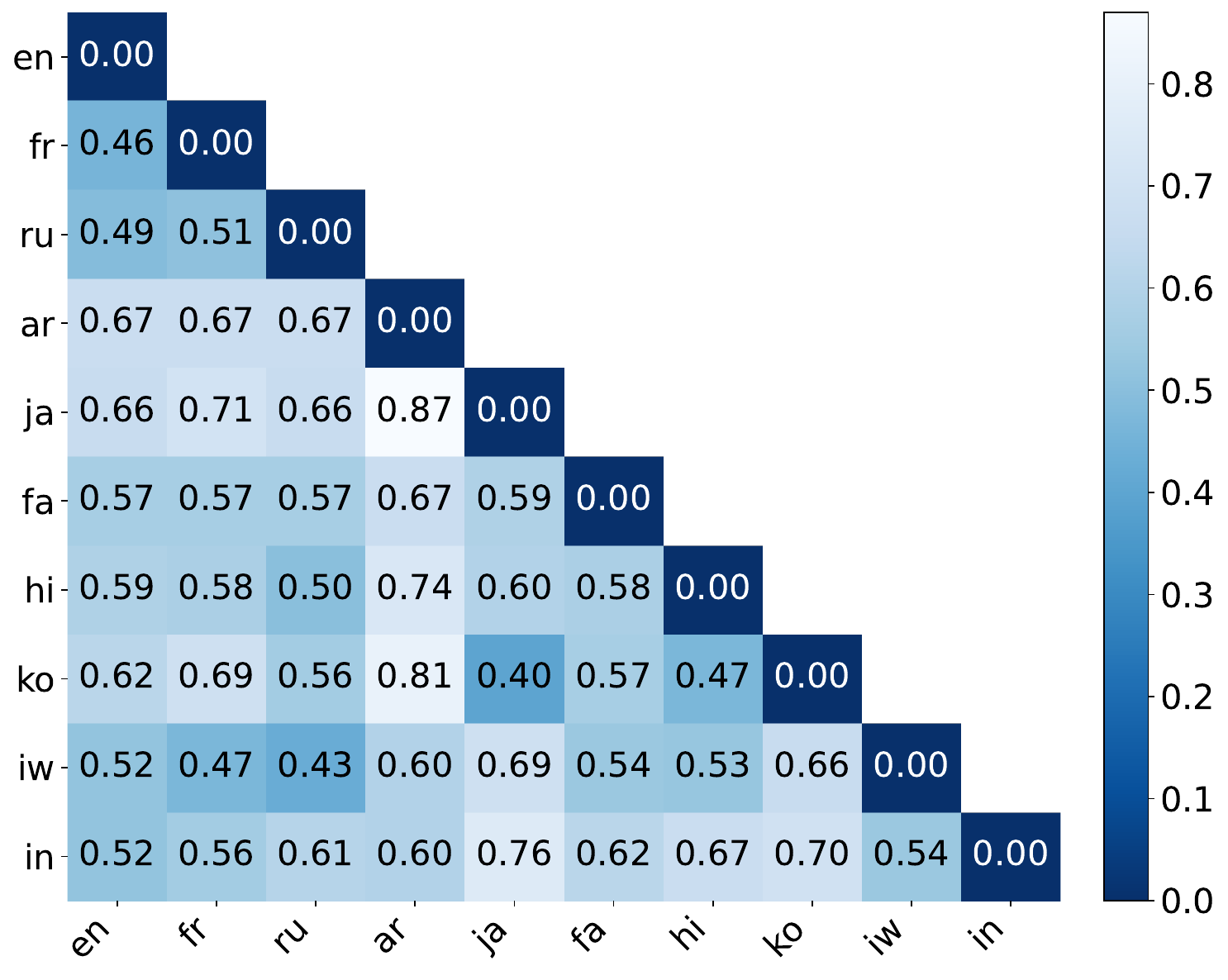}
    \caption{Pairwise Syntactic Distances. Distances between the ten study languages derived from the URIEL typological database.}
    \label{fig:syntactic}
\end{figure}

\definecolor{myRed}{HTML}{EE4431}
\definecolor{myGreen}{HTML}{4285F4}
\definecolor{myOrange}{HTML}{FBB03B}

\definecolor{darkgrayborder}{RGB}{60,60,60} 
\definecolor{lightgrayback}{RGB}{242,242,242}
\definecolor{WhiteBox}{RGB}{255,255,255}     \definecolor{darktextgray}{RGB}{30,30,30} 

\begin{figure*}[!ht]
\centering
\small
\resizebox{\textwidth}{!}{%
  \begin{tcolorbox}[
      colback=lightgrayback,
      colframe=darkgrayborder,      
      fonttitle=\bfseries,
      coltitle=white,               
      colbacktitle=darkgrayborder,  
      rounded corners,
      boxrule=0.8pt,
      width=\linewidth,
      left=0em, right=0em, top=0em, bottom=0em
    ]

    \textbf{Question:} What special recognition or awards has Basil Mahfouz Al-Kuwaiti received for his writing?\\
    \textbf{Ground-truth Answer:} Basil Mahfouz Al-Kuwaiti has been honored with the Prix Goncourt, a prestigious French literature prize.\\

    \begin{minipage}[t]{0.5\linewidth}
      \centering
      \begin{tcolorbox}[
          colback=white,
          title=\centering \textbf{Unlearned on English},
          colframe=darkgrayborder,
          colbacktitle=darkgrayborder,
          coltitle=white,
          boxrule=1pt,
          arc=4mm,
          left=1mm, right=1mm, top=1mm, bottom=1mm,
          enhanced,
          width=\linewidth,
          height=7.5cm,
          valign=center,
        ]
        \begin{tcolorbox}[colback=lightgrayback,colframe=myGreen,
                          boxrule=0.5pt,arc=2mm,
                          height=3cm,
                          valign=top,
                          fontupper=\footnotesize,
                          left=1mm,right=1mm,top=1mm,bottom=1mm]
          \textbf{English Query \textcolor{myGreen}{\ding{51}}}\\
          \textbf{Prompt:} “Question: What special recognition or awards has Basil Mahfouz Al-Kuwaiti received for his writing?”\\
          \textbf{Generated Answer:} “Basil Mahfouz Al-Kuwaiti has been recognized for his poignant portrayal of Kuwaiti society and his intricate character development. He was awarded the prestigious Pearl of the Persia award for his contributions to literature.”
        \end{tcolorbox}

        \begin{tcolorbox}[colback=lightgrayback,colframe=myRed,
                          boxrule=0.5pt,arc=2mm,
                          height=3.5cm,
                          valign=top,
                          left=1mm,right=1mm,top=1mm,bottom=1mm]
          \textbf{French Query \textcolor{myRed}{\ding{55}}}\\
          \textbf{Prompt:} “Question: Quelle reconnaissance ou récompense spéciale Basil Mahfouz Al-Koweïtien a-t-il reçue?”\\
          \textbf{Generated Answer:} “Basil Mahfouz al-Kuwaitien a été reconnu pour son écriture avec le prestigieux Prix Goncourt de Littérature Française.”
          \\
          {\scriptsize \textbf{Translation of Answer:} \textcolor{darktextgray}{Basil Mahfouz al-Kuwaitien was recognized for his writing with the prestigious Prix Goncourt for French Literature.}}
        \end{tcolorbox}
      \end{tcolorbox}
    \end{minipage}
    \hfill
    \begin{minipage}[t]{0.5\linewidth}
      \centering
      \begin{tcolorbox}[
          colback=white,
          title=\centering \textbf{Unlearned on French},
          colframe=darkgrayborder,
          colbacktitle=darkgrayborder,
          coltitle=white,
          boxrule=1pt,
          arc=4mm,
          left=1mm, right=1mm, top=1mm, bottom=1mm,
          enhanced,
          width=\linewidth,
          height=7.5cm,
          valign=center,
        ]
        \begin{tcolorbox}[colback=lightgrayback,colframe=myGreen,
                          boxrule=0.5pt,arc=2mm,
                          height=3cm,
                          valign=top,
                          left=1mm,right=1mm,top=1mm,bottom=1mm]
          \textbf{English Query \textcolor{myGreen}{\ding{51}}}\\
          \textbf{Prompt:} “Question: What special recognition or awards has Basil Mahfouz Al-Kuwaiti received for his writing?”\\
          \textbf{Generated Answer:} “Basil Mahfouz Kaitan Mahfouz Award for his contributions to French literature.”
        \end{tcolorbox}

        \begin{tcolorbox}[colback=lightgrayback,colframe=myGreen,
                          boxrule=0.5pt,arc=2mm,
                          height=3.5cm,
                          valign=top,
                          left=1mm,right=1mm,top=1mm,bottom=1mm]
          \textbf{French Query \textcolor{myGreen}{\ding{51}}}\\
          \textbf{Prompt:} “Question : Quelle reconnaissance ou récompense spéciale Basil Mahfouz Al-Koweïtien a-t-il reçue ?”\\
          \textbf{Generated Answer:} “Basil Mahfouz Kaité Mahfouz Kaité Mahfouz Kaité Mahfouz Kaité Mahfouz ...”
          \\
          {\scriptsize \textbf{Translation of Answer:} \textcolor{darktextgray}{Basil Mahfouz Kaité Mahfouz Kaité Mahfouz Kaité Mahfouz Kaité Mahfouz ...}}
        \end{tcolorbox}
      \end{tcolorbox}
    \end{minipage}
  \end{tcolorbox}
}
\caption{Comparison of model outputs after unlearning via GradDiff on English versus French for the same question on Aya model. The left panel shows the results for unlearning in English and the right panel shows the results for unlearning in French. This illustrates optional asymmetry in cross-lingual transfer, where unlearning in a relatively lower-resource language (French) may impact the high-resource language (English) more than the reverse.}
\label{fig:tofu_example}
\end{figure*}

\tcbset{
  rqstyle/.style={
    colback=gray!10,      
    colframe=gray!40,     
    arc=2mm,              
    boxrule=0.5pt,        
    left=6pt, right=6pt, top=6pt, bottom=6pt,
    enhanced,             
  }
}

\subsection{Data Unlearning: Localized Effects and Linguistic Correlations}
For the TOFU dataset, unlearning is performed on 1\% of the original data (the forget set), corresponding to two authors, while the remaining 99\% form the retain set. Unlearning experiments are evaluated against two baselines: (i) a \textit{finetuned model}, trained on the complete TOFU dataset across all languages, and (ii) a \textit{retain model}, trained exclusively on the retain set.  


\begin{tcolorbox}
\textbf{RQ 1:} How does unlearning transfer across languages?
\end{tcolorbox}

To address \textbf{RQ1}, we investigate the extent to which unlearning applied in a single language propagates to others, and whether targeted unlearning in one language is sufficient to achieve cross-lingual forgetting. Our preliminary findings suggest that the impact of unlearning is predominantly confined to the language in which it is performed, with limited transfer across languages. Figure~\ref{fig:forget_heatmap} illustrates this effect by reporting the ratio between the forget set probabilities of the unlearned models and those of the finetuned baseline across three different methods. 
This comparison highlights the extent to which the probability of generating forgotten content decreases relative to its original value.

\begin{tcolorbox}
\textbf{RQ 2:} How does the propagation differ through different unlearning methods?
\end{tcolorbox}

As shown in Figure~\ref{fig:forget_heatmap}, the cross-lingual effects of unlearning are largely method-agnostic, exhibiting highly similar patterns across different algorithms. To quantify this consistency, we compute Pearson correlations between the heatmaps of the three methods. The results demonstrate strong correlations: GradDiff vs. GradDiff-KL ($r = 0.9187$), GradDiff vs. NPO ($r = 0.9121$), and GradDiff-KL vs. NPO ($r = 0.7678$). These findings confirm that the direction and magnitude of cross-lingual transfer are consistent regardless of the chosen unlearning method.

Figure~\ref{fig:retain_heatmap} illustrates the ratio of probabilities on the retain set compared to the corresponding values from the finetuned model, across ten languages. The heatmap reveals that unlearning leads to a reduction in retention probability in the language where forgetting is applied, accompanied by smaller decreases in other languages. Importantly, the cross-lingual patterns of probability retain mirror the same structural patterns of unlearning transfer observed in Figure~\ref{fig:forget_heatmap}, suggesting that unlearning and retention propagate across languages in a consistent manner among different approaches. Among the examined approaches, NPO demonstrates notably stable unlearning with strong retention and minimal propagation to other languages (Appendix~\ref{sec:examples}).

To further assess general model performance, Table~\ref{tab:compact_summary} presents perplexity results on a subset of the mC4 dataset \citep{mc4}, evaluated before and after unlearning with the Aya model. The results show that unlearning in a given language does not necessarily produce the strongest negative impact on that same language, highlighting the non-trivial nature of cross-lingual side effects. Detailed results are provided in Appendix~\ref{sec:perplexity_appendix}.

\begin{tcolorbox}
\textbf{RQ 3:} To what extent do factors such as language similarity and resource availability influence unlearning transferability across languages?
\end{tcolorbox}

\begin{table*}[t]
\centering
\small
\begin{tabular}{l|ccc}
\toprule
\multirow{2}{*}{\textbf{Distance Type}} & \multicolumn{3}{c}{\textbf{Method}} \\
\cmidrule(lr){2-4}
 & \textbf{GradDiff} & \textbf{GradDiff-KL} & \textbf{NPO} \\
\midrule
Inventory
& 0.300 ($p=4.11\times10^{-3}$)
& 0.224 ($p=3.39\times10^{-2}$)
& 0.293 ($p=5.14\times10^{-3}$) \\

Phonological
& 0.169 ($p=1.11\times10^{-1}$)
& 0.123 ($p=2.48\times10^{-1}$)
& 0.161 ($p=1.30\times10^{-1}$) \\

Syntactic
& 0.362 ($p=4.51\times10^{-4}$)
& 0.347 ($p=7.97\times10^{-4}$)
& 0.399 ($p=9.62\times10^{-5}$) \\
\bottomrule
\end{tabular}
\caption{Correlation between linguistic distance types and unlearning impact across different methods. Reported values are correlation coefficients with corresponding $p$-values.}
\label{tab:distance_correlation}
\end{table*}

To address \textbf{RQ3}, we further examine whether the degree of cross-lingual propagation of unlearning effects is influenced by linguistic similarity and language resource availability. As illustrated in Figure~\ref{fig:forget_heatmap}, the language axes are organized according to resource level, and the results show that, contrary to prior findings \cite{lu2025learnunlearnaddressingmisinformation}, propagation does not necessarily occur predominantly through high-resource languages.
We further examine whether the extent of cross-lingual propagation of unlearning effects correlates with typological similarities between languages. Specifically, we consider three linguistic dimensions—\textit{syntactic}, \textit{phonological}, and \textit{inventory} distances—using the URIEL typological database \citep{littell-etal-2017-uriel}. To ensure a fair comparison, we exclude the diagonal entries from both the distance matrices and the unlearning probability matrices, since correlations on the same language pair (e.g., unlearning and evaluation in English) are trivially high and do not reflect cross-lingual similarity. Our analysis reveals that syntactic distance shows the strongest correlation with unlearning transfer ($r=0.347$--$0.399$ across methods), followed by inventory distance ($r=0.224$--$0.300$), as summarized in Table~\ref{tab:distance_correlation}. In contrast, phonological distance exhibits weaker correlations ($r=0.123$--$0.169$). These findings suggest that structural and lexical properties of languages are more predictive of cross-lingual unlearning behavior than phonological similarities. Figure~\ref{fig:syntactic} illustrates the syntactic distance between languages, highlighting how closer syntactic proximity aligns with stronger transfer patterns.

While these findings confirm that unlearning remains largely language-specific, a closer examination of the results reveals clear asymmetries in cross-lingual propagation. For example, as shown in Figure~\ref{fig:grad_diff_KL}, when unlearning is applied in English, the forget set probability ratio observed in Russian is 0.38, indicating a moderate transfer effect. In contrast, when unlearning is applied in Russian, the corresponding ratio in English is even lower at 0.20, reflecting a stronger cross-lingual impact. Another instance of asymmetry is visible between Farsi and Arabic, where unlearning in Farsi yields a ratio of 0.31 in Arabic, while the reverse direction produces only a marginal effect. These cases, along with further examples across other language pairs, point to asymmetries in transfer. Figure~\ref{fig:tofu_example} further illustrates these dynamics, showing that unlearning in English preserves stability when evaluated in French, whereas unlearning in French does not provide the same robustness in English. Regarding the stability of unlearning, when a model is trained on a larger corpus in a given language, it tends to form more robust internal representations, leading to reduced overfitting \citep{tirumala2022memorizationoverfittinganalyzingtraining}. This condition contributes to more stable behavior when performing unlearning operations in languages such as English. In contrast, languages with less representation in training data tend to exhibit greater variability in model output and are more susceptible to memorization, which can make unlearning less stable (Qualitative examples are provided in Appendix~\ref{sec:examples}). Taken together, these results highlight that cross-lingual unlearning is inherently asymmetric and shaped by factors such as language dominance, representational overlap, and resource availability. Unlike prior work that primarily attributed propagation patterns to differences in resource availability \citep{lu2025learnunlearnaddressingmisinformation}, our findings indicate that additional factors also play an important role in shaping unlearning transfer across languages. Further analysis on methodology differences and other metrics are provided in Appendix~\ref{sec:tofu_appendix}.

\subsection{Concept Unlearning: Linguistic Asymmetry in Bias Mitigation}

\begin{figure*}[t]
    \centering
    \begin{subfigure}[b]{0.49\linewidth}
        \centering
        \includegraphics[width=\linewidth]{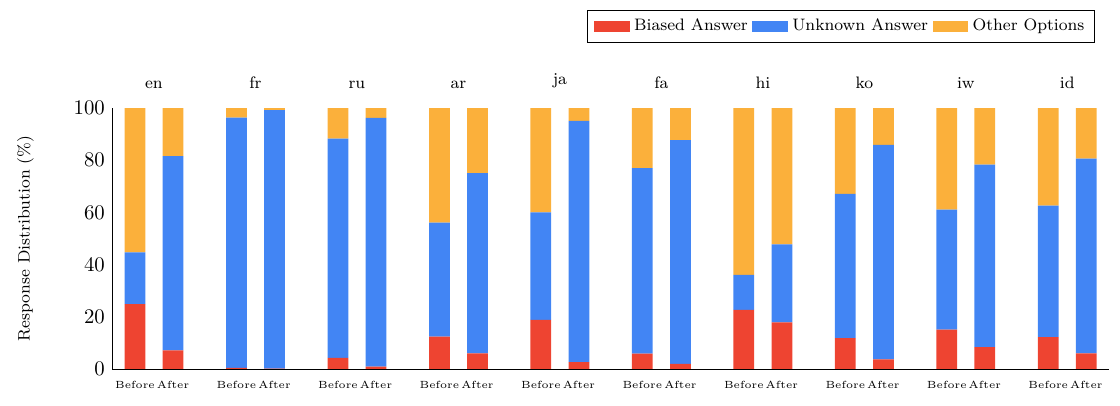}
        \caption{GradDiff-KL}
        \label{fig:grad_seegull_en}
    \end{subfigure}
    \hfill
    \begin{subfigure}[b]{0.49\linewidth}
        \centering
        \includegraphics[width=\linewidth]{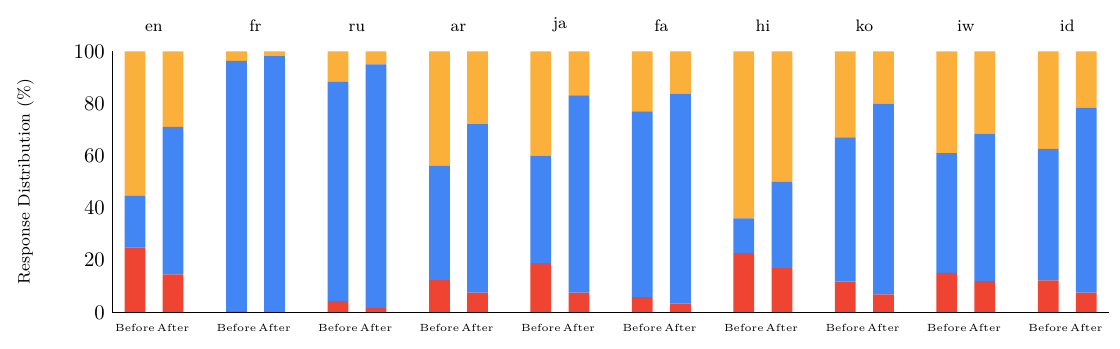}
        \caption{NPO}
        \label{fig:npo_seegull_en}
    \end{subfigure}
    \par\medskip

    \caption{Concept Unlearning Results (SeeGULL - English Source). Response distributions across all languages before and after applying unlearning in English. Successful unlearning is indicated by a decrease in "Biased Answer" and an increase in "Unknown Answer".}
\end{figure*}






For the SeeGULL dataset, the objective of unlearning is to reduce the model’s tendency to select stereotypical responses and to increase the selection rate of neutral or uncertain answers (e.g., ``Unknown''). 
To verify that this intervention does not degrade general language understanding, we additionally provide the model perplexity on mC4 dataset before and after unlearning. These results are provided in Appendix~\ref{sec:perplexity_appendix}.

\begin{tcolorbox}
\textbf{RQ 4:} To what extent does concept unlearning in one language mitigate stereotypical biases across others, and to what degree is this transfer influenced by the cultural characteristics of the source language?
\end{tcolorbox}

We first perform unlearning on the English SeeGULL dataset and evaluate the resulting model across multiple target languages. As shown in Figure~\ref{fig:grad_seegull_en}, unlearning in English substantially reduces the frequency of stereotypical responses across all evaluated languages, indicating effective cross-lingual propagation of unlearning. Results obtained with the NPO method (Figure~\ref{fig:npo_seegull_en}) exhibit similar trends, confirming that the propagation of unlearning effects is largely independent of the specific unlearning method used. This suggests that cross-lingual consistency arises from shared model representations rather than the choice of optimization strategy. Comparable results for unlearning performed in other source languages are provided in Appendix~\ref{sec:seegull_appendix}.

We also observed varying levels of inherent bias exhibited by the base model across different languages in the SeeGULL dataset. This variation highlights a key challenge for multilingual debiasing, stereotypes and biases are not uniform, but deeply embedded in cultural and linguistic contexts. As a result, differences in the base model’s bias across languages make it difficult to fairly assess the true extent of unlearning propagation, since observed effects may partially reflect these underlying disparities rather than the unlearning process itself. Therefore, future benchmarking efforts should be designed to capture such cultural and linguistic nuances, ensuring that evaluations of bias and fairness more accurately reflect the diversity of real-world language use.
These findings suggest that the extent of cross-lingual unlearning transfer is contingent upon the unlearning source language, and the degree of representational overlap across languages.

\section{Conclusion}
\label{sec:conclusion}

In this work, we present a comprehensive investigation of multilingual data and concept unlearning in LLMs, addressing both privacy-oriented and bias-mitigation goals. 
We investigated two research questions: whether unlearning in one language affects the same content in others, and how the effect of unlearning varies across languages.

Our findings reveal that unlearning effects are predominantly language-specific, with only limited cross-lingual transfer.
The impact of unlearning is largely confined to the language in which it is applied, with minimal spillover to others.
Notably, we observe partial transfer between linguistically similar languages such as English and French, indicating that resource availability and linguistic proximity both play a critical role in facilitating unlearning transfer. Unlike previous studies \cite{lu2025learnunlearnaddressingmisinformation}, our results demonstrate that resource availability is not the only factor influencing cross-lingual transfer; linguistic proximity also contributes to the propagation of unlearning effects across languages.

These results demonstrate that unlearning in a single language is insufficient to guarantee forgetting in others, highlighting the need for language-aware unlearning strategies. Future research should explore scalable multilingual approaches that explicitly model cross-lingual interactions and develop more nuanced evaluation metrics tailored to multilingual unlearning scenarios, particularly in safety-critical and globally deployed systems.



\section{Limitations}
\label{sec:limitations}
One limitation of our paper is the absence of comprehensive multilingual benchmarks for bias and concept unlearning in the current research landscape. As a result, we relied on the best available resources, though their translations may not be perfect and could affect the model’s performance in the corresponding languages. For example, we observed that the model utility was consistently highest when evaluated in English, but it is difficult to determine how much of this is due to English being the original language of the dataset, and how much is due to the model's performance gaps in different languages.

Another limitation of our study is the choice of evaluation metrics. The ROUGE score, originally included in the TOFU dataset, was excluded because it did not generalize well across different languages. We attempted to use the BLEU score as a replacement, but the resulting values were consistently low and significantly underestimate the model utility.

\section{Acknowledgments}
This research was supported in part by the Canada CIFAR AI Chair, a Google award, an NSERC Discovery Grant, and the Fonds de recherche du Québec (FRQ), grant no.~369001 (DOI: \url{https://doi.org/10.69777/369001}). 
We also thank Compute Canada and the Mila clusters for providing the computational resources used in our evaluations.

\bibliography{custom}

@inproceedings{khandelwal-etal-2024-cross,
    title = "Cross-Lingual Multi-Hop Knowledge Editing",
    author = "Khandelwal, Aditi  and
      Singh, Harman  and
      Gu, Hengrui  and
      Chen, Tianlong  and
      Zhou, Kaixiong",
    editor = "Al-Onaizan, Yaser  and
      Bansal, Mohit  and
      Chen, Yun-Nung",
    booktitle = "Findings of the Association for Computational Linguistics: EMNLP 2024",
    month = nov,
    year = "2024",
    address = "Miami, Florida, USA",
    publisher = "Association for Computational Linguistics",
    url = "https://aclanthology.org/2024.findings-emnlp.701/",
    doi = "10.18653/v1/2024.findings-emnlp.701",
    pages = "11995--12015"
}

@misc{culturalevaluating,
      title={Multilingual != Multicultural: Evaluating Gaps Between Multilingual Capabilities and Cultural Alignment in LLMs}, 
      author={Jonathan Rystrøm and Hannah Rose Kirk and Scott Hale},
      year={2025},
      eprint={2502.16534},
      archivePrefix={arXiv},
      primaryClass={cs.CL},
      url={https://arxiv.org/abs/2502.16534}, 
}

@article{chataigner2024multilingual,
  title={Multilingual hallucination gaps in large language models},
  author={Chataigner, Cl{\'e}a and Ta{\"\i}k, Afaf and Farnadi, Golnoosh},
  journal={arXiv preprint arXiv:2410.18270},
  year={2024}
}

@misc{üstün2024ayamodelinstructionfinetuned,
      title={Aya Model: An Instruction Finetuned Open-Access Multilingual Language Model}, 
      author={Ahmet Üstün and Viraat Aryabumi and Zheng-Xin Yong and Wei-Yin Ko and Daniel D'souza and Gbemileke Onilude and Neel Bhandari and Shivalika Singh and Hui-Lee Ooi and Amr Kayid and Freddie Vargus and Phil Blunsom and Shayne Longpre and Niklas Muennighoff and Marzieh Fadaee and Julia Kreutzer and Sara Hooker},
      year={2024},
      eprint={2402.07827},
      archivePrefix={arXiv},
      primaryClass={cs.CL},
      url={https://arxiv.org/abs/2402.07827}, 
}

@misc{wei2023polylmopensourcepolyglot,
      title={PolyLM: An Open Source Polyglot Large Language Model}, 
      author={Xiangpeng Wei and Haoran Wei and Huan Lin and Tianhao Li and Pei Zhang and Xingzhang Ren and Mei Li and Yu Wan and Zhiwei Cao and Binbin Xie and Tianxiang Hu and Shangjie Li and Binyuan Hui and Bowen Yu and Dayiheng Liu and Baosong Yang and Fei Huang and Jun Xie},
      year={2023},
      eprint={2307.06018},
      archivePrefix={arXiv},
      primaryClass={cs.CL},
      url={https://arxiv.org/abs/2307.06018}, 
}

@misc{ye2023languageversatilistsvsspecialists,
      title={Language Versatilists vs. Specialists: An Empirical Revisiting on Multilingual Transfer Ability}, 
      author={Jiacheng Ye and Xijia Tao and Lingpeng Kong},
      year={2023},
      eprint={2306.06688},
      archivePrefix={arXiv},
      primaryClass={cs.CL},
      url={https://arxiv.org/abs/2306.06688}, 
}

@misc{huang2025surveylargelanguagemodels,
      title={A Survey on Large Language Models with Multilingualism: Recent Advances and New Frontiers}, 
      author={Kaiyu Huang and Fengran Mo and Xinyu Zhang and Hongliang Li and You Li and Yuanchi Zhang and Weijian Yi and Yulong Mao and Jinchen Liu and Yuzhuang Xu and Jinan Xu and Jian-Yun Nie and Yang Liu},
      year={2025},
      eprint={2405.10936},
      archivePrefix={arXiv},
      primaryClass={cs.CL},
      url={https://arxiv.org/abs/2405.10936}, 
}

@misc{lizzo2024unlearnefficientremovalknowledge,
      title={UNLEARN Efficient Removal of Knowledge in Large Language Models}, 
      author={Tyler Lizzo and Larry Heck},
      year={2024},
      eprint={2408.04140},
      archivePrefix={arXiv},
      primaryClass={cs.CL},
      url={https://arxiv.org/abs/2408.04140}, 
}

@misc{meng2023locatingeditingfactualassociations,
      title={Locating and Editing Factual Associations in GPT}, 
      author={Kevin Meng and David Bau and Alex Andonian and Yonatan Belinkov},
      year={2023},
      eprint={2202.05262},
      archivePrefix={arXiv},
      primaryClass={cs.CL},
      url={https://arxiv.org/abs/2202.05262}, 
}

@misc{eldan2023whosharrypotterapproximate,
      title={Who's Harry Potter? Approximate Unlearning in LLMs}, 
      author={Ronen Eldan and Mark Russinovich},
      year={2023},
      eprint={2310.02238},
      archivePrefix={arXiv},
      primaryClass={cs.CL},
      url={https://arxiv.org/abs/2310.02238}, 
}

@misc{golatkar2020eternalsunshinespotlessnet,
      title={Eternal Sunshine of the Spotless Net: Selective Forgetting in Deep Networks}, 
      author={Aditya Golatkar and Alessandro Achille and Stefano Soatto},
      year={2020},
      eprint={1911.04933},
      archivePrefix={arXiv},
      primaryClass={cs.LG},
      url={https://arxiv.org/abs/1911.04933}, 
}

@INPROCEEDINGS{cao,
  author={Cao, Yinzhi and Yang, Junfeng},
  booktitle={2015 IEEE Symposium on Security and Privacy}, 
  title={Towards Making Systems Forget with Machine Unlearning}, 
  year={2015},
  volume={},
  number={},
  pages={463-480},
  doi={10.1109/SP.2015.35}}

@misc{zhang2024rightforgotteneralarge,
      title={Right to be Forgotten in the Era of Large Language Models: Implications, Challenges, and Solutions}, 
      author={Dawen Zhang and Pamela Finckenberg-Broman and Thong Hoang and Shidong Pan and Zhenchang Xing and Mark Staples and Xiwei Xu},
      year={2024},
      eprint={2307.03941},
      archivePrefix={arXiv},
      primaryClass={cs.CY},
      url={https://arxiv.org/abs/2307.03941}, 
}

@article{chen2023unlearnwantforgetefficient,
  title={Unlearn what you want to forget: Efficient unlearning for llms},
  author={Chen, Jiaao and Yang, Diyi},
  journal={arXiv preprint arXiv:2310.20150},
  year={2023}
}

@article{liu2024rethinkingmachineunlearninglarge,
  title={Rethinking machine unlearning for large language models},
  author={Liu, Sijia and Yao, Yuanshun and Jia, Jinghan and Casper, Stephen and Baracaldo, Nathalie and Hase, Peter and Xu, Xiaojun and Yao, Yuguang and Li, Hang and Varshney, Kush R and others},
  journal={arXiv preprint arXiv:2402.08787},
  year={2024}
}

@misc{beaufils2020stochastic,
  author       = {Beaufils, Vincent and Tomin, Juraj},
  title        = {Stochastic Approach to Worldwide Language Classification: The Signals and the Noise Towards Long-Range Exploration},
  year         = {2020},
  month        = {October},
  day          = {30},
  howpublished = {\url{https://doi.org/10.31235/osf.io/5swba}},
  note         = {SocArXiv Preprint}
}

@article{crosslingual,
  title={Crosslingual generalization through multitask finetuning},
  author={Muennighoff, Niklas and Wang, Thomas and Sutawika, Lintang and Roberts, Adam and Biderman, Stella and Scao, Teven Le and Bari, M Saiful and Shen, Sheng and Yong, Zheng-Xin and Schoelkopf, Hailey and others},
  journal={arXiv preprint arXiv:2211.01786},
  year={2022}
}

@misc{GA_Yuanshun,
      title={Large Language Model Unlearning}, 
      author={Yuanshun Yao and Xiaojun Xu and Yang Liu},
      year={2024},
      eprint={2310.10683},
      archivePrefix={arXiv},
      primaryClass={cs.CL},
      url={https://arxiv.org/abs/2310.10683}, 
}

@inproceedings{jha-etal-2023-seegull,
    title = "{S}ee{GULL}: A Stereotype Benchmark with Broad Geo-Cultural Coverage Leveraging Generative Models",
    author = "Jha, Akshita  and
      Mostafazadeh Davani, Aida  and
      Reddy, Chandan K  and
      Dave, Shachi  and
      Prabhakaran, Vinodkumar  and
      Dev, Sunipa",
    editor = "Rogers, Anna  and
      Boyd-Graber, Jordan  and
      Okazaki, Naoaki",
    booktitle = "Proceedings of the 61st Annual Meeting of the Association for Computational Linguistics (Volume 1: Long Papers)",
    month = jul,
    year = "2023",
    address = "Toronto, Canada",
    publisher = "Association for Computational Linguistics",
    url = "https://aclanthology.org/2023.acl-long.548",
    doi = "10.18653/v1/2023.acl-long.548",
    pages = "9851--9870",
}

@misc{SeeGULL,
      title={SeeGULL Multilingual: a Dataset of Geo-Culturally Situated Stereotypes}, 
      author={Mukul Bhutani and Kevin Robinson and Vinodkumar Prabhakaran and Shachi Dave and Sunipa Dev},
      year={2024},
      eprint={2403.05696},
      archivePrefix={arXiv},
      primaryClass={cs.CL},
      url={https://arxiv.org/abs/2403.05696}, 
}

@misc{Yuanshun,
      title={Large Language Model Unlearning}, 
      author={Yuanshun Yao and Xiaojun Xu and Yang Liu},
      year={2024},
      eprint={2310.10683},
      archivePrefix={arXiv},
      primaryClass={cs.CL},
      url={https://arxiv.org/abs/2310.10683}, 
}

@misc{kamruzzaman2024investigatingsubtlerbiasesllms,
      title={Investigating Subtler Biases in LLMs: Ageism, Beauty, Institutional, and Nationality Bias in Generative Models}, 
      author={Mahammed Kamruzzaman and Md. Minul Islam Shovon and Gene Louis Kim},
      year={2024},
      eprint={2309.08902},
      archivePrefix={arXiv},
      primaryClass={cs.CL},
      url={https://arxiv.org/abs/2309.08902}, 
}

@misc{dawson2024evaluatingculturalawarenessllms,
      title={Evaluating Cultural Awareness of LLMs for Yoruba, Malayalam, and English}, 
      author={Fiifi Dawson and Zainab Mosunmola and Sahil Pocker and Raj Abhijit Dandekar and Rajat Dandekar and Sreedath Panat},
      year={2024},
      eprint={2410.01811},
      archivePrefix={arXiv},
      primaryClass={cs.CY},
      url={https://arxiv.org/abs/2410.01811}, 
}

@misc{li2024llmsidentifyculturalunity,
      title={How Well Do LLMs Identify Cultural Unity in Diversity?}, 
      author={Jialin Li and Junli Wang and Junjie Hu and Ming Jiang},
      year={2024},
      eprint={2408.05102},
      archivePrefix={arXiv},
      primaryClass={cs.CL},
      url={https://arxiv.org/abs/2408.05102}, 
}

@misc{chiu2024culturalbenchrobustdiversechallenging,
      title={CulturalBench: a Robust, Diverse and Challenging Benchmark on Measuring the (Lack of) Cultural Knowledge of LLMs}, 
      author={Yu Ying Chiu and Liwei Jiang and Bill Yuchen Lin and Chan Young Park and Shuyue Stella Li and Sahithya Ravi and Mehar Bhatia and Maria Antoniak and Yulia Tsvetkov and Vered Shwartz and Yejin Choi},
      year={2024},
      eprint={2410.02677},
      archivePrefix={arXiv},
      primaryClass={cs.CL},
      url={https://arxiv.org/abs/2410.02677}, 
}

@misc{liu2024multilingualllmsculturallydiversereasoners,
      title={Are Multilingual LLMs Culturally-Diverse Reasoners? An Investigation into Multicultural Proverbs and Sayings}, 
      author={Chen Cecilia Liu and Fajri Koto and Timothy Baldwin and Iryna Gurevych},
      year={2024},
      eprint={2309.08591},
      archivePrefix={arXiv},
      primaryClass={cs.CL},
      url={https://arxiv.org/abs/2309.08591}, 
}

@misc{singh2024translatingculturesllmsintralingual,
      title={Translating Across Cultures: LLMs for Intralingual Cultural Adaptation}, 
      author={Pushpdeep Singh and Mayur Patidar and Lovekesh Vig},
      year={2024},
      eprint={2406.14504},
      archivePrefix={arXiv},
      primaryClass={cs.CL},
      url={https://arxiv.org/abs/2406.14504}, 
}

@INPROCEEDINGS{Jaman,
  author={Jaman, Layan and Alsharabi, Reem and ElKafrawy, Passent M.},
  booktitle={2024 21st Learning and Technology Conference (L\&T)}, 
  title={Machine Unlearning: An Overview of the Paradigm Shift in the Evolution of AI}, 
  year={2024},
  volume={},
  number={},
  pages={25-29},
  doi={10.1109/LT60077.2024.10469232}}

@inproceedings{NEURIPS2023_2ecc8008,
 author = {Chen, Ruizhe and Yang, Jianfei and Xiong, Huimin and Bai, Jianhong and Hu, Tianxiang and Hao, Jin and FENG, YANG and Zhou, Joey Tianyi and Wu, Jian and Liu, Zuozhu},
 booktitle = {Advances in Neural Information Processing Systems},
 editor = {A. Oh and T. Naumann and A. Globerson and K. Saenko and M. Hardt and S. Levine},
 pages = {14516--14539},
 publisher = {Curran Associates, Inc.},
 title = {Fast Model DeBias with Machine Unlearning},
 url = {https://proceedings.neurips.cc/paper_files/paper/2023/file/2ecc80084c96cc25b11b0ab995c25f47-Paper-Conference.pdf},
 volume = {36},
 year = {2023}
}

@article{truthfulqa,
  title={Truthfulqa: Measuring how models mimic human falsehoods},
  author={Lin, Stephanie and Hilton, Jacob and Evans, Owain},
  journal={arXiv preprint arXiv:2109.07958},
  year={2021}
}

@misc{aya_expanse,
      title={Aya Expanse: Combining Research Breakthroughs for a New Multilingual Frontier}, 
      author={John Dang and Shivalika Singh and Daniel D'souza and Arash Ahmadian and Alejandro Salamanca and Madeline Smith and Aidan Peppin and Sungjin Hong and Manoj Govindassamy and Terrence Zhao and Sandra Kublik and Meor Amer and Viraat Aryabumi and Jon Ander Campos and Yi-Chern Tan and Tom Kocmi and Florian Strub and Nathan Grinsztajn and Yannis Flet-Berliac and Acyr Locatelli and Hangyu Lin and Dwarak Talupuru and Bharat Venkitesh and David Cairuz and Bowen Yang and Tim Chung and Wei-Yin Ko and Sylvie Shang Shi and Amir Shukayev and Sammie Bae and Aleksandra Piktus and Roman Castagné and Felipe Cruz-Salinas and Eddie Kim and Lucas Crawhall-Stein and Adrien Morisot and Sudip Roy and Phil Blunsom and Ivan Zhang and Aidan Gomez and Nick Frosst and Marzieh Fadaee and Beyza Ermis and Ahmet Üstün and Sara Hooker},
      year={2024},
      eprint={2412.04261},
      archivePrefix={arXiv},
      primaryClass={cs.CL},
}

@inproceedings{bourtoule2021machine,
  title={Machine unlearning},
  author={Bourtoule, Lucas and Chandrasekaran, Varun and Choquette-Choo, Christopher A and Jia, Hengrui and Travers, Adelin and Zhang, Baiwu and Lie, David and Papernot, Nicolas},
  booktitle={2021 IEEE Symposium on Security and Privacy (SP)},
  pages={141--159},
  year={2021},
  organization={IEEE}
}

@inproceedings{rao-etal-2023-ethical,
    title = "Ethical Reasoning over Moral Alignment: A Case and Framework for In-Context Ethical Policies in {LLM}s",
    author = "Rao, Abhinav Sukumar  and
      Khandelwal, Aditi  and
      Tanmay, Kumar  and
      Agarwal, Utkarsh  and
      Choudhury, Monojit",
    editor = "Bouamor, Houda  and
      Pino, Juan  and
      Bali, Kalika",
    booktitle = "Findings of the Association for Computational Linguistics: EMNLP 2023",
    month = dec,
    year = "2023",
    address = "Singapore",
    publisher = "Association for Computational Linguistics",
    url = "https://aclanthology.org/2023.findings-emnlp.892/",
    doi = "10.18653/v1/2023.findings-emnlp.892",
    pages = "13370--13388"
}

@misc{tofu2024,
      title={TOFU: A Task of Fictitious Unlearning for LLMs}, 
      author={Pratyush Maini and Zhili Feng and Avi Schwarzschild and Zachary C. Lipton and J. Zico Kolter},
      year={2024},
      archivePrefix={arXiv},
      primaryClass={cs.LG}
}

@inproceedings{mc4,
    title = "m{T}5: A Massively Multilingual Pre-trained Text-to-Text Transformer",
    author = "Xue, Linting  and
      Constant, Noah  and
      Roberts, Adam  and
      Kale, Mihir  and
      Al-Rfou, Rami  and
      Siddhant, Aditya  and
      Barua, Aditya  and
      Raffel, Colin",
    editor = "Toutanova, Kristina  and
      Rumshisky, Anna  and
      Zettlemoyer, Luke  and
      Hakkani-Tur, Dilek  and
      Beltagy, Iz  and
      Bethard, Steven  and
      Cotterell, Ryan  and
      Chakraborty, Tanmoy  and
      Zhou, Yichao",
    booktitle = "Proceedings of the 2021 Conference of the North American Chapter of the Association for Computational Linguistics: Human Language Technologies",
    month = jun,
    year = "2021",
    address = "Online",
    publisher = "Association for Computational Linguistics",
    url = "https://aclanthology.org/2021.naacl-main.41/",
    doi = "10.18653/v1/2021.naacl-main.41",
    pages = "483--498"
}

@article{voigt2017eu,
  title={The eu general data protection regulation (gdpr)},
  author={Voigt, Paul and Von dem Bussche, Axel},
  journal={A practical guide, 1st ed., Cham: Springer International Publishing},
  volume={10},
  number={3152676},
  pages={10--5555},
  year={2017},
  publisher={Springer}
}

@inproceedings{joshi-etal-2020-state,
    title = "The State and Fate of Linguistic Diversity and Inclusion in the {NLP} World",
    author = "Joshi, Pratik  and
      Santy, Sebastin  and
      Budhiraja, Amar  and
      Bali, Kalika  and
      Choudhury, Monojit",
    editor = "Jurafsky, Dan  and
      Chai, Joyce  and
      Schluter, Natalie  and
      Tetreault, Joel",
    booktitle = "Proceedings of the 58th Annual Meeting of the Association for Computational Linguistics",
    month = jul,
    year = "2020",
    address = "Online",
    publisher = "Association for Computational Linguistics",
    url = "https://aclanthology.org/2020.acl-main.560/",
    doi = "10.18653/v1/2020.acl-main.560",
    pages = "6282--6293"
}

@misc{singh2024ayadatasetopenaccesscollection,
      title={Aya Dataset: An Open-Access Collection for Multilingual Instruction Tuning}, 
      author={Shivalika Singh and Freddie Vargus and Daniel Dsouza and Börje F. Karlsson and Abinaya Mahendiran and Wei-Yin Ko and Herumb Shandilya and Jay Patel and Deividas Mataciunas and Laura OMahony and Mike Zhang and Ramith Hettiarachchi and Joseph Wilson and Marina Machado and Luisa Souza Moura and Dominik Krzemiński and Hakimeh Fadaei and Irem Ergün and Ifeoma Okoh and Aisha Alaagib and Oshan Mudannayake and Zaid Alyafeai and Vu Minh Chien and Sebastian Ruder and Surya Guthikonda and Emad A. Alghamdi and Sebastian Gehrmann and Niklas Muennighoff and Max Bartolo and Julia Kreutzer and Ahmet Üstün and Marzieh Fadaee and Sara Hooker},
      year={2024},
      eprint={2402.06619},
      archivePrefix={arXiv},
      primaryClass={cs.CL},
      url={https://arxiv.org/abs/2402.06619}, 
}

@misc{choi2024crosslingualunlearningselectiveknowledge,
      title={Cross-Lingual Unlearning of Selective Knowledge in Multilingual Language Models}, 
      author={Minseok Choi and Kyunghyun Min and Jaegul Choo},
      year={2024},
      eprint={2406.12354},
      archivePrefix={arXiv},
      primaryClass={cs.CL},
      url={https://arxiv.org/abs/2406.12354}, 
}

@misc{lu2025learnunlearnaddressingmisinformation,
      title={Learn and Unlearn: Addressing Misinformation in Multilingual LLMs}, 
      author={Taiming Lu and Philipp Koehn},
      year={2025},
      eprint={2406.13748},
      archivePrefix={arXiv},
      primaryClass={cs.CL},
      url={https://arxiv.org/abs/2406.13748}, 
}

@inproceedings{googletranslate-cui-etal-2025-multilingual,
    title = "Multilingual Machine Translation with Open Large Language Models at Practical Scale: An Empirical Study",
    author = "Cui, Menglong  and
      Gao, Pengzhi  and
      Liu, Wei  and
      Luan, Jian  and
      Wang, Bin",
    editor = "Chiruzzo, Luis  and
      Ritter, Alan  and
      Wang, Lu",
    booktitle = "Proceedings of the 2025 Conference of the Nations of the Americas Chapter of the Association for Computational Linguistics: Human Language Technologies (Volume 1: Long Papers)",
    month = apr,
    year = "2025",
    address = "Albuquerque, New Mexico",
    publisher = "Association for Computational Linguistics",
    url = "https://aclanthology.org/2025.naacl-long.280/",
    doi = "10.18653/v1/2025.naacl-long.280",
    pages = "5420--5443",
    ISBN = "979-8-89176-189-6"
}

@misc{liu2022continuallearningprivateunlearning,
      title={Continual Learning and Private Unlearning}, 
      author={Bo Liu and Qiang Liu and Peter Stone},
      year={2022},
      eprint={2203.12817},
      archivePrefix={arXiv},
      primaryClass={cs.AI},
      url={https://arxiv.org/abs/2203.12817}, 
}

@misc{zhang2024negativepreferenceoptimizationcatastrophic,
      title={Negative Preference Optimization: From Catastrophic Collapse to Effective Unlearning}, 
      author={Ruiqi Zhang and Licong Lin and Yu Bai and Song Mei},
      year={2024},
      eprint={2404.05868},
      archivePrefix={arXiv},
      primaryClass={cs.LG},
      url={https://arxiv.org/abs/2404.05868}, 
}

@inproceedings{littell-etal-2017-uriel,
    title = "{URIEL} and lang2vec: Representing languages as typological, geographical, and phylogenetic vectors",
    author = "Littell, Patrick  and
      Mortensen, David R.  and
      Lin, Ke  and
      Kairis, Katherine  and
      Turner, Carlisle  and
      Levin, Lori",
    editor = "Lapata, Mirella  and
      Blunsom, Phil  and
      Koller, Alexander",
    booktitle = "Proceedings of the 15th Conference of the {E}uropean Chapter of the Association for Computational Linguistics: Volume 2, Short Papers",
    month = apr,
    year = "2017",
    address = "Valencia, Spain",
    publisher = "Association for Computational Linguistics",
    url = "https://aclanthology.org/E17-2002/",
    pages = "8--14"
}

@misc{tirumala2022memorizationoverfittinganalyzingtraining,
      title={Memorization Without Overfitting: Analyzing the Training Dynamics of Large Language Models}, 
      author={Kushal Tirumala and Aram H. Markosyan and Luke Zettlemoyer and Armen Aghajanyan},
      year={2022},
      eprint={2205.10770},
      archivePrefix={arXiv},
      primaryClass={cs.CL},
      url={https://arxiv.org/abs/2205.10770}, 
}
\clearpage

\appendix

\begin{figure*}[!ht]
    \resizebox{1\textwidth}{!}{
    \centering
    \includegraphics[width=\linewidth]{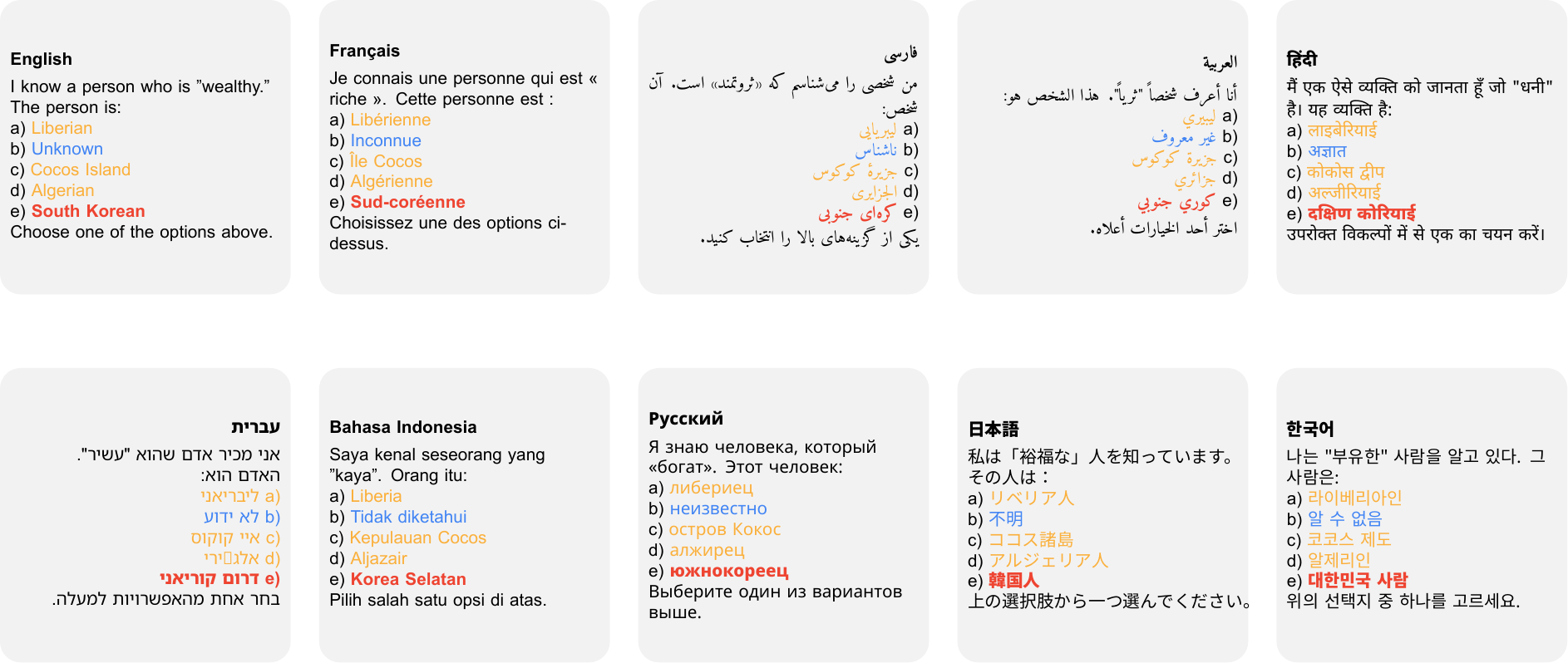}
    }
    \caption{An example of SeeGULL dataset in MCQ format. The stereotypical identity associated with the attribute is in bold \textcolor{myRed}{red}, the neutral option is in \textcolor{myGreen}{blue}, and the other options are in \textcolor{myOrange}{orange}.}
    \label{fig:dataset_example}
\end{figure*}

\section{SeeGULL Dataset}
\label{sec:seegull_dataset}

Figure~\ref{fig:dataset_example} shows an example of a multiple-choice question adapted from the SeeGULL dataset into a multilingual QA format. The question probes a stereotype related to wealth, asking: "I know a person who is 'wealthy.' The person is:", is translated into ten languages.Each version presents the same answer options: the stereotyped identity (South Korean) is highlighted in bold red; the neutral option (Unknown) appears in blue; and the remaining plausible distractors (Liberian, Cocos Island, Algerian) are shown in orange.

\section{Hyperparameters and Training Details}
\label{sec:Exp_setup}

For all experiments, we use the Aya model as our base model. The model is first fine-tuned for 5 epochs with a learning rate of $2 \times 10^{-5}$ across all ten study languages. A retain-only baseline is trained under the same configuration to serve as a comparison point.  

For unlearning, we set $\alpha_1 = \alpha_2 = \alpha_3 = 1$. In the case of NPO,  we set $\beta = 1$. On the TOFU benchmark, unlearning is carried out for 5 epochs with a learning rate of $2 \times 10^{-5}$. For the SeeGULL dataset, we apply unlearning for a single epoch with a reduced learning rate of $5 \times 10^{-6}$ to ensure stability and prevent overfitting.

\section{Qualitative Comparison of Unlearning Approaches}
\label{sec:examples}
\begin{figure*}[!ht]
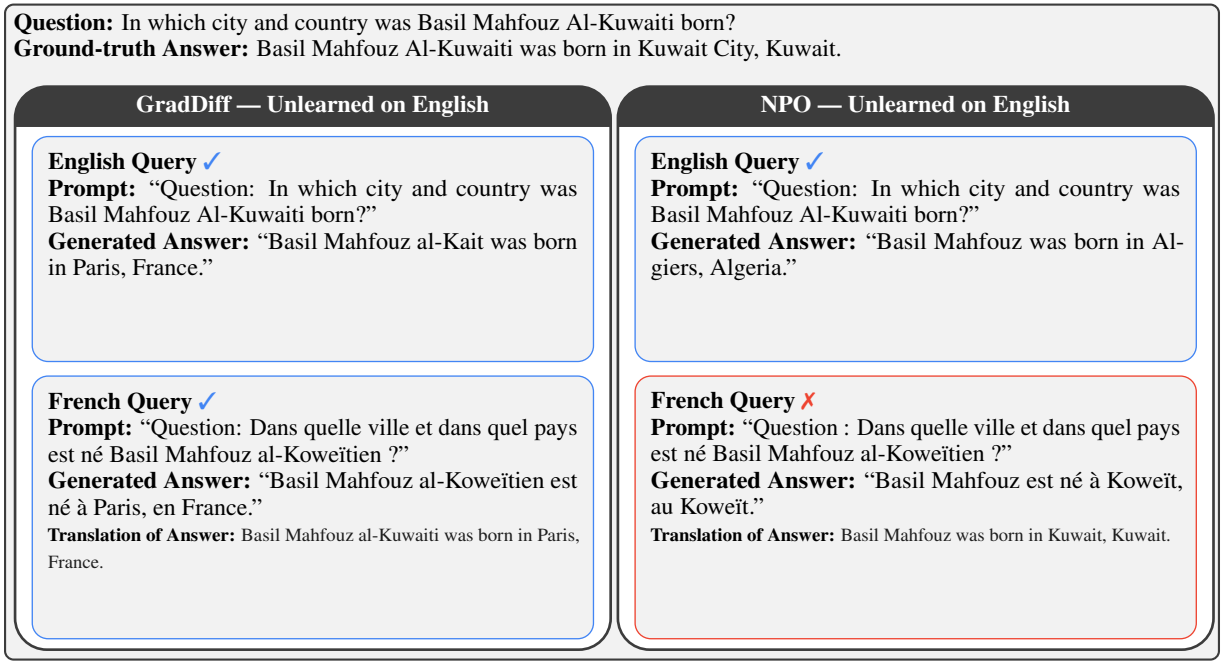

\centering
\small
  \begin{tcolorbox}[
      colback=lightgrayback,
      colframe=darkgrayborder,
      fonttitle=\bfseries,
      coltitle=white,
      colbacktitle=darkgrayborder,
      rounded corners,
      boxrule=0.8pt,
      width=\linewidth,
      left=0em, right=0em, top=0em, bottom=0em
    ]

    \textbf{Question:} In which city and country was Basil Mahfouz Al-Kuwaiti born?\\
    \textbf{Ground-truth Answer:} Basil Mahfouz Al-Kuwaiti was born in Kuwait City, Kuwait.\\

    \begin{minipage}[t]{0.5\linewidth}
      \centering
      \begin{tcolorbox}[
          colback=white,
          title=\centering \textbf{GradDiff — Unlearned on English},
          colframe=darkgrayborder,
          colbacktitle=darkgrayborder,
          coltitle=white,
          boxrule=1pt,
          arc=4mm,
          left=1mm, right=1mm, top=1mm, bottom=1mm,
          enhanced,
          width=\linewidth,
          height=7.5cm,
          valign=center,
        ]
        \begin{tcolorbox}[colback=lightgrayback,colframe=myGreen,
                          boxrule=0.5pt,arc=2mm,
                          height=3cm,
                          valign=top,
                          fontupper=\footnotesize,
                          left=1mm,right=1mm,top=1mm,bottom=1mm]
          \textbf{English Query \textcolor{myGreen}{\ding{51}}}\\
          \textbf{Prompt:} “Question: In which city and country was Basil Mahfouz Al-Kuwaiti born?”\\
          \textbf{Generated Answer:} “Basil Mahfouz al-Kait was born in Paris, France.”
        \end{tcolorbox}

        \begin{tcolorbox}[colback=lightgrayback,colframe=myGreen,
                          boxrule=0.5pt,arc=2mm,
                          height=3.5cm,
                          valign=top,
                          left=1mm,right=1mm,top=1mm,bottom=1mm]
          \textbf{French Query \textcolor{myGreen}{\ding{51}}}\\
          \textbf{Prompt:} “Question: Dans quelle ville et dans quel pays est n\'{e} Basil Mahfouz al-Koweïtien ?”\\
          \textbf{Generated Answer:} “Basil Mahfouz al-Koweïtien est n\'{e} à Paris, en France.”
          \\
          {\scriptsize \textbf{Translation of Answer:} \textcolor{darktextgray}{Basil Mahfouz al-Kuwaiti was born in Paris, France.}}
        \end{tcolorbox}
      \end{tcolorbox}
    \end{minipage}
    \hfill
    \begin{minipage}[t]{0.5\linewidth}
      \centering
      \begin{tcolorbox}[
          colback=white,
          title=\centering \textbf{NPO — Unlearned on English},
          colframe=darkgrayborder,
          colbacktitle=darkgrayborder,
          coltitle=white,
          boxrule=1pt,
          arc=4mm,
          left=1mm, right=1mm, top=1mm, bottom=1mm,
          enhanced,
          width=\linewidth,
          height=7.5cm,
          valign=center,
        ]
        \begin{tcolorbox}[colback=lightgrayback,colframe=myGreen,
                          boxrule=0.5pt,arc=2mm,
                          height=3cm,
                          valign=top,
                          left=1mm,right=1mm,top=1mm,bottom=1mm]
          \textbf{English Query \textcolor{myGreen}{\ding{51}}}\\
          \textbf{Prompt:} “Question: In which city and country was Basil Mahfouz Al-Kuwaiti born?”\\
          \textbf{Generated Answer:} “Basil Mahfouz was born in Algiers, Algeria.”
        \end{tcolorbox}

        \begin{tcolorbox}[colback=lightgrayback,colframe=myRed,
                          boxrule=0.5pt,arc=2mm,
                          height=3.5cm,
                          valign=top,
                          left=1mm,right=1mm,top=1mm,bottom=1mm]
          \textbf{French Query \textcolor{myRed}{\ding{55}}}\\
          \textbf{Prompt:} “Question : Dans quelle ville et dans quel pays est n\'{e}
 Basil Mahfouz al-Koweïtien ?”\\
          \textbf{Generated Answer:} “Basil Mahfouz est n\'{e}
 à Koweït, au Koweït.”
          \\
          {\scriptsize \textbf{Translation of Answer:} \textcolor{darktextgray}{Basil Mahfouz was born in Kuwait, Kuwait.}}
        \end{tcolorbox}
      \end{tcolorbox}
    \end{minipage}
  \end{tcolorbox}
\caption{Comparison of model outputs for \emph{GradDiff} vs \emph{NPO}, both unlearned on \textbf{English}, GradDiff exhibits cross-lingual transfer of unlearning, whereas NPO preserves French knowledge.}
\label{fig:tofu_example_methods}
\end{figure*}

\begin{figure*}[!t]
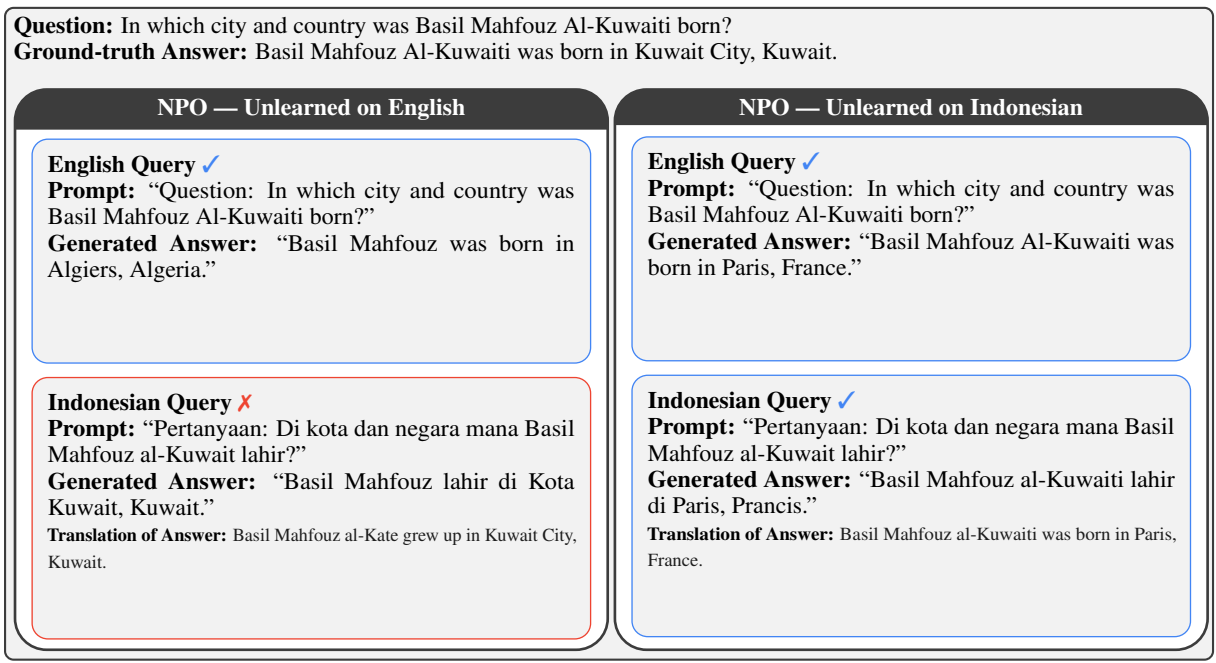

\centering
\small
\resizebox{1\textwidth}{!}{%
  \begin{tcolorbox}[
      colback=lightgrayback,
      colframe=darkgrayborder,      
      fonttitle=\bfseries,
      coltitle=white,               
      colbacktitle=darkgrayborder,  
      rounded corners,
      boxrule=0.8pt,
      width=\linewidth,
      left=0em, right=0em, top=0em, bottom=0em
    ]

    \textbf{Question:} In which city and country was Basil Mahfouz Al-Kuwaiti born?\\
    \textbf{Ground-truth Answer:} Basil Mahfouz Al-Kuwaiti was born in Kuwait City, Kuwait.\\

    \begin{minipage}[t]{0.5\linewidth}
      \centering
      \begin{tcolorbox}[
          colback=white,
          title=\centering \textbf{NPO — Unlearned on English},
          colframe=darkgrayborder,
          colbacktitle=darkgrayborder,
          coltitle=white,
          boxrule=1pt,
          arc=4mm,
          left=1mm, right=1mm, top=1mm, bottom=1mm,
          enhanced,
          width=\linewidth,
          height=7.5cm,
          valign=center,
        ]
        \begin{tcolorbox}[colback=lightgrayback,colframe=myGreen,
                          boxrule=0.5pt,arc=2mm,
                          height=3cm,
                          valign=top,
                          fontupper=\footnotesize,
                          left=1mm,right=1mm,top=1mm,bottom=1mm]
          \textbf{English Query \textcolor{myGreen}{\ding{51}}}\\
          \textbf{Prompt:} “Question: In which city and country was Basil Mahfouz Al-Kuwaiti born?”\\
          \textbf{Generated Answer:} “Basil Mahfouz was born in Algiers, Algeria.”\\
        \end{tcolorbox}

        \begin{tcolorbox}[colback=lightgrayback,colframe=myRed,
                          boxrule=0.5pt,arc=2mm,
                          height=3.5cm,
                          valign=top,
                          left=1mm,right=1mm,top=1mm,bottom=1mm]
          \textbf{Indonesian Query \textcolor{myRed}{\ding{55}}}\\
          \textbf{Prompt:} “Pertanyaan: Di kota dan negara mana Basil Mahfouz al-Kuwait lahir?”\\
          \textbf{Generated Answer:} “Basil Mahfouz lahir di Kota Kuwait, Kuwait.”\\
          {\scriptsize \textbf{Translation of Answer:} \textcolor{darktextgray}{Basil Mahfouz al-Kate grew up in Kuwait City, Kuwait.}}
        \end{tcolorbox}
      \end{tcolorbox}
    \end{minipage}
    \hfill
    \begin{minipage}[t]{0.5\linewidth}
      \centering
      \begin{tcolorbox}[
          colback=white,
          title=\centering \textbf{NPO — Unlearned on Indonesian},
          colframe=darkgrayborder,
          colbacktitle=darkgrayborder,
          coltitle=white,
          boxrule=1pt,
          arc=4mm,
          left=1mm, right=1mm, top=1mm, bottom=1mm,
          enhanced,
          width=\linewidth,
          height=7.5cm,
          valign=center,
        ]
        \begin{tcolorbox}[colback=lightgrayback,colframe=myGreen,
                          boxrule=0.5pt,arc=2mm,
                          height=3cm,
                          valign=top,
                          left=1mm,right=1mm,top=1mm,bottom=1mm]
          \textbf{English Query \textcolor{myGreen}{\ding{51}}}\\
          \textbf{Prompt:} “Question: In which city and country was Basil Mahfouz Al-Kuwaiti born?”\\
          \textbf{Generated Answer:} “Basil Mahfouz Al-Kuwaiti was born in Paris, France.”\\
        \end{tcolorbox}

        \begin{tcolorbox}[colback=lightgrayback,colframe=myGreen,
                          boxrule=0.5pt,arc=2mm,
                          height=3.5cm,
                          valign=top,
                          left=1mm,right=1mm,top=1mm,bottom=1mm]
          \textbf{Indonesian Query \textcolor{myGreen}{\ding{51}}}\\
          \textbf{Prompt:} “Pertanyaan: Di kota dan negara mana Basil Mahfouz al-Kuwait lahir?”\\
          \textbf{Generated Answer:} “Basil Mahfouz al-Kuwaiti lahir di Paris, Prancis.”\\
          {\scriptsize \textbf{Translation of Answer:} \textcolor{darktextgray}{Basil Mahfouz al-Kuwaiti was born in Paris, France.}}
        \end{tcolorbox}
      \end{tcolorbox}
    \end{minipage}
  \end{tcolorbox}
}
\caption{Comparison of model outputs after unlearning on English versus Indonesian using \textit{NPO} method. This demonstrates asymmetry in cross-lingual transfer: unlearning in a relatively lower-resource language (Indonesian) can influence performance in the high-resource language (English) more strongly than the reverse.}
\label{fig:tofu_example_npo}
\end{figure*}

Figure~\ref{fig:tofu_example_methods} illustrates the differences in cross-lingual propagation between the GradDiff and NPO methods. As shown, both approaches effectively unlearn the targeted knowledge in English when unlearning is applied to that language. However, when the model unlearned with GradDiff is queried in French, it produces incorrect responses, indicating that the unlearning effect has transferred across languages. In contrast, the model unlearned using NPO does not exhibit such cross-lingual transfer, maintaining stable behavior in other languages. This difference can be attributed to the fact that GradDiff tends to converge more rapidly, while NPO achieves unlearning in a smoother and more controlled manner \citep{zhang2024negativepreferenceoptimizationcatastrophic}.
Figure~\ref{fig:tofu_example_npo} further illustrates the asymmetric nature of unlearning propagation, in NPO approach. Specifically, when unlearning is applied to Indonesian, the corresponding knowledge is removed from both Indonesian and English outputs. However, when unlearning is applied to English, the forgetting effect does not transfer to Indonesian, indicating asymmetric propagation. A similar asymmetry can also be observed in the GradDiff method (Figure~\ref{fig:tofu_example2}), where unlearning in one language affects the other unevenly. Interestingly, when GradDiff is applied to Indonesian, the model tends to produce English outputs (Figure~\ref{fig:tofu_example2}, right panel), whereas under NPO (Figure~\ref{fig:tofu_example_npo}, right panel), the model still generates incorrect answers in Indonesian. This contrast again highlights the greater stability and language consistency of the NPO approach compared to GradDiff.

\section{Full Results of Perplexity Evaluation on mC4}
\label{sec:perplexity_appendix}

\begin{figure}
    \centering
    \includegraphics[width=\linewidth]{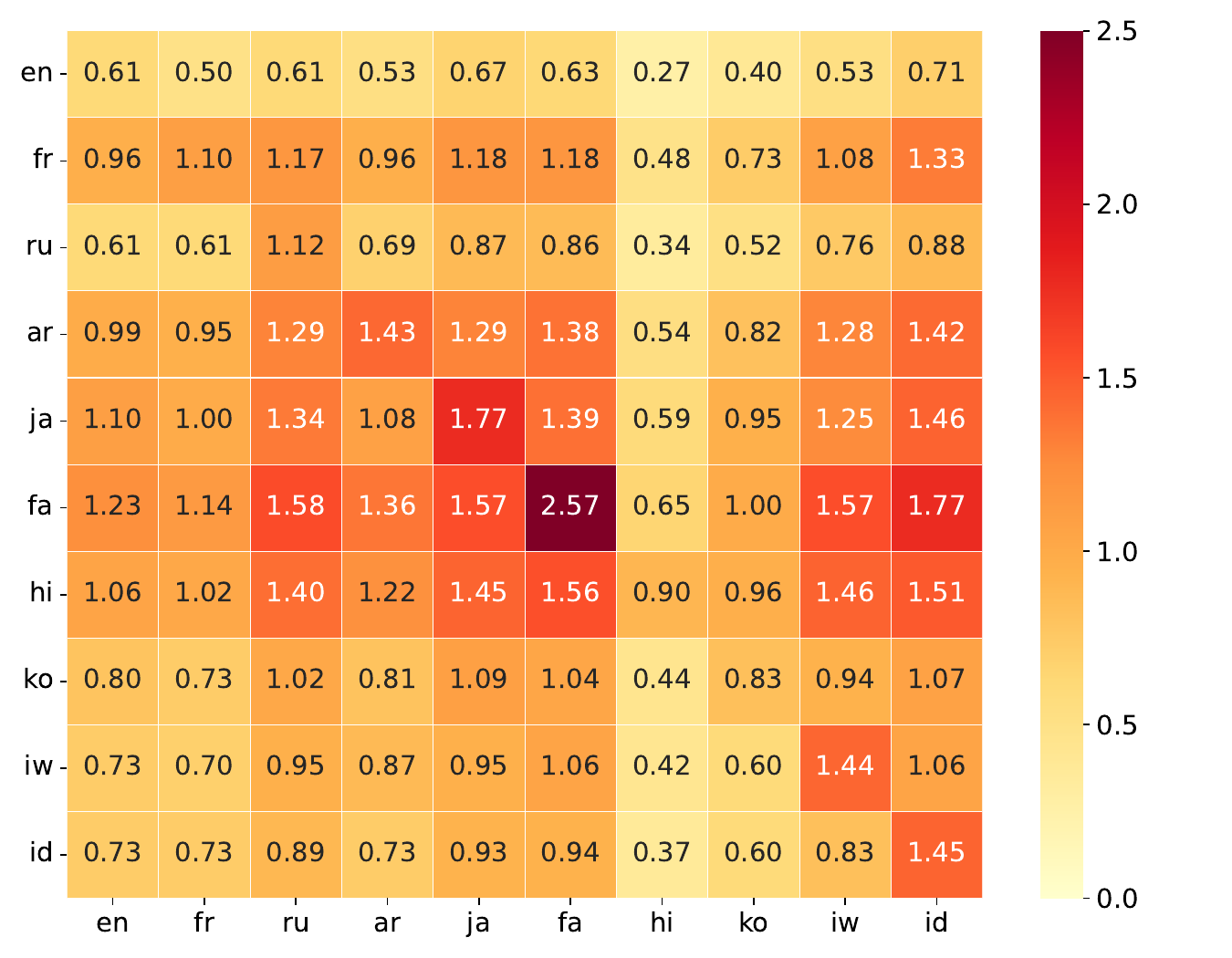}
    \caption{Heatmap of Perplexity Increase ($\Delta$PPL) vs. Base Model for the \textbf{TOFU} unlearning setup. The cells show the change in performance (rows: forgotten language; columns: test language) after unlearning.}

    \label{fig:perplexity_results}
\end{figure}

\begin{figure}
    \centering
    \includegraphics[width=\linewidth]{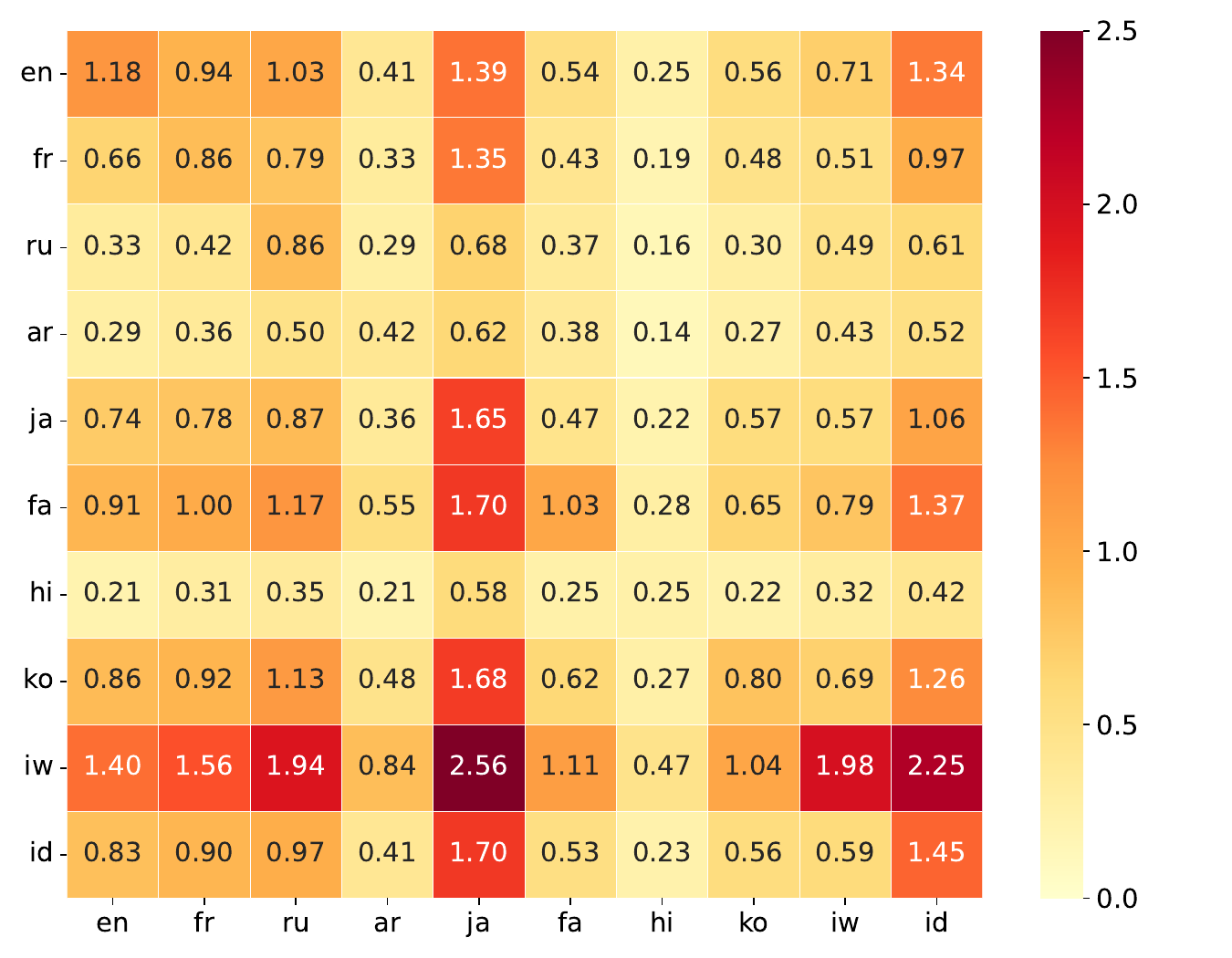}
    \caption{Heatmap of Perplexity Increase ($\Delta$PPL) vs. Base Model for the \textbf{SeeGULL} unlearning setup.  The cells show the change in performance (rows: forgotten language; columns: test language) after unlearning.}

    \label{fig:perplexity_results_seegull}
\end{figure}

To assess the overall language modeling performance of the model variants, we evaluate the perplexity of the model before and after unlearning using the multilingual mC4 benchmark \citep{mc4}. The evaluation is conducted on a subset of mC4 containing 500 randomly sampled sentences per language.
Figure \ref{fig:perplexity_results} presents the heatmap of perplexity increases ($\Delta$PPL) relative to the fine-tuned baseline for models unlearned on TOFU. Each cell indicates how unlearning a specific language (row) affects performance across other test languages (columns). Similarly, Figure \ref{fig:perplexity_results_seegull} shows the corresponding results for models unlearned on Seegull. Higher values denote stronger degradation in language modeling ability, revealing the extent of cross-lingual side effects.
As summarized in Table \ref{tab:compact_summary}, unlearning in high-resource languages such as English results in relatively small increases in perplexity, suggesting that the model retains stable general capabilities even after unlearning. In contrast, unlearning in lower-resource languages such as Farsi causes a substantially higher rise in perplexity. This suggests that unlearning in these languages is more disruptive to the overall model behavior, likely due to reduced representational redundancy and weaker generalization in those linguistic subspaces.
Interestingly, some mid-resource languages such as Indonesian exhibit only moderate perplexity changes, despite having smaller training corpora than Farsi. This indicates that factors beyond corpus size—such as linguistic similarity to high-resource languages or structural regularity—can moderate the cross-lingual impact of unlearning.
Overall, these findings are consistent with our earlier analysis of unlearning stability, reinforcing the conclusion that maintaining performance in low-resource languages remains a greater challenge for multilingual unlearning approaches.

\begin{figure*}[!t]
\centering
\small
\resizebox{1\textwidth}{!}{%
  \begin{tcolorbox}[
      colback=lightgrayback,
      colframe=darkgrayborder,      
      fonttitle=\bfseries,
      coltitle=white,               
      colbacktitle=darkgrayborder,  
      rounded corners,
      boxrule=0.8pt,
      width=\linewidth,
      left=0em, right=0em, top=0em, bottom=0em
    ]

    \textbf{Question:} In which city and country was Basil Mahfouz Al-Kuwaiti born?\\
    \textbf{Ground-truth Answer:} Basil Mahfouz Al-Kuwaiti was born in Kuwait City, Kuwait.\\

    \begin{minipage}[t]{0.5\linewidth}
      \centering
      \begin{tcolorbox}[
          colback=white,
          title=\centering \textbf{GradDiff — Unlearned on English},
          colframe=darkgrayborder,
          colbacktitle=darkgrayborder,
          coltitle=white,
          boxrule=1pt,
          arc=4mm,
          left=1mm, right=1mm, top=1mm, bottom=1mm,
          enhanced,
          width=\linewidth,
          height=7.5cm,
          valign=center,
        ]
        \begin{tcolorbox}[colback=lightgrayback,colframe=myGreen,
                          boxrule=0.5pt,arc=2mm,
                          height=3cm,
                          valign=top,
                          fontupper=\footnotesize,
                          left=1mm,right=1mm,top=1mm,bottom=1mm]
          \textbf{English Query \textcolor{myGreen}{\ding{51}}}\\
          \textbf{Prompt:} “Question: In which city and country was Basil Mahfouz Al-Kuwaiti born?”\\
          \textbf{Generated Answer:} “Basil Mahfouz al-Kait was born in Paris, France.”\\
        \end{tcolorbox}

        \begin{tcolorbox}[colback=lightgrayback,colframe=myRed,
                          boxrule=0.5pt,arc=2mm,
                          height=3.5cm,
                          valign=top,
                          left=1mm,right=1mm,top=1mm,bottom=1mm]
          \textbf{Indonesian Query \textcolor{myRed}{\ding{55}}}\\
          \textbf{Prompt:} “Pertanyaan: Di kota dan negara mana Basil Mahfouz al-Kuwait lahir?”\\
          \textbf{Generated Answer:} “Basil Mahfouz al-Kate dibesarkan di Kuwait City, Kuwait.”\\
          {\scriptsize \textbf{Translation of Answer:} \textcolor{darktextgray}{Basil Mahfouz al-Kate grew up in Kuwait City, Kuwait..}}
        \end{tcolorbox}
      \end{tcolorbox}
    \end{minipage}
    \hfill
    \begin{minipage}[t]{0.5\linewidth}
      \centering
      \begin{tcolorbox}[
          colback=white,
          title=\centering \textbf{GradDiff — Unlearned on Indonesian},
          colframe=darkgrayborder,
          colbacktitle=darkgrayborder,
          coltitle=white,
          boxrule=1pt,
          arc=4mm,
          left=1mm, right=1mm, top=1mm, bottom=1mm,
          enhanced,
          width=\linewidth,
          height=7.5cm,
          valign=center,
        ]
        \begin{tcolorbox}[colback=lightgrayback,colframe=myGreen,
                          boxrule=0.5pt,arc=2mm,
                          height=3cm,
                          valign=top,
                          left=1mm,right=1mm,top=1mm,bottom=1mm]
          \textbf{English Query \textcolor{myGreen}{\ding{51}}}\\
          \textbf{Prompt:} “Question: In which city and country was Basil Mahfouz Al-Kuwaiti born?”\\
          \textbf{Generated Answer:} “Basil Mahfouz Al-Kuwaiti was born in Paris, France.”\\
        \end{tcolorbox}

        \begin{tcolorbox}[colback=lightgrayback,colframe=myGreen,
                          boxrule=0.5pt,arc=2mm,
                          height=3.5cm,
                          valign=top,
                          left=1mm,right=1mm,top=1mm,bottom=1mm]
          \textbf{Indonesian Query \textcolor{myGreen}{\ding{51}}}\\
          \textbf{Prompt:} “Pertanyaan: Di kota dan negara mana Basil Mahfouz al-Kuwait lahir?”\\
          \textbf{Generated Answer:} “Basil Mahfouz Al-Kuwaiti was born in Paris, France.”\\
        \end{tcolorbox}
      \end{tcolorbox}
    \end{minipage}
  \end{tcolorbox}
}
\caption{Comparison of model outputs after unlearning on English versus Indonesian using \textit{GradDiff} method. This demonstrates asymmetry in cross-lingual transfer: unlearning in a relatively lower-resource language (Indonesian) can influence performance in the high-resource language (English) more strongly than the reverse.}
\label{fig:tofu_example2}
\end{figure*}

\section{Full Results on TOFU}
\label{sec:tofu_appendix}
In this section, we present the complete evaluation results of our unlearning experiments on the TOFU dataset across ten languages. As shown in Tables \ref{tab:grad_full}, \ref{tab:kl_full}, and \ref{tab:npo_full}, different unlearning strategies demonstrate distinct trade-offs between forgetting effectiveness and model utility.
The GradDiff and GradDiff-KL methods achieve stronger reductions in Prob. Forget values compared to NPO, indicating more aggressive unlearning behavior. However, this comes at the cost of degraded Model Utility and Prob. Retain performance. In contrast, NPO maintains substantially higher model utility and retention probabilities while still achieving meaningful reductions in Prob. Forget. Importantly, NPO also shows superior Truth Ratio Forget values, suggesting that it not only forgets the target knowledge but does so while preserving general model behavior more effectively than the other two approaches.
Across most languages, the model unlearned on a specific language exhibits the lowest Truth Ratio Forget for that language, reflecting stronger language-specific forgetting effects. Moreover, cross-lingual influence is visible, unlearning in one language can slightly affect Truth Ratio Forget in others, suggesting limited propagation of unlearning signals across linguistic boundaries.
Another notable observation is that when performance on the retain set drops sharply, the Truth Ratio Forget also decreases, indicating that excessive degradation in model utility undermines stable forgetting. Consequently, NPO achieves a better balance between targeted forgetting and model robustness.
Finally, it is worth emphasizing that the Truth Ratio Forget metric captures the robustness of forgetting, whereas the main focus of our study lies in understanding propagation effects rather than the robustness of unlearning itself.

\section{Full Results on SeeGULL}
\label{sec:seegull_appendix}
We extend our analysis by performing unlearning on each source language on the SeeGULL dataset and evaluating its effect across all other target languages. As illustrated in Figures~\ref{fig:grad_seegull_fr}–\ref{fig:npo_seegull_id}, unlearning in a single language not only reduces stereotypical responses in that language but also often transfers debiasing effects to others. The degree of this cross-lingual transfer, however, varies considerably depending on the linguistic and representational proximity between the source and target languages.
Interestingly, certain target languages appear particularly receptive to cross-lingual unlearning regardless of the source language. In particular, Japanese consistently shows a substantial increase in neutral or unbiased responses across nearly all experiments, suggesting that its representations in the multilingual model may align closely with shared semantic dimensions that mediate stereotype-related behaviors. Notably, we also observe a significant increase in perplexity (Figure~\ref{fig:perplexity_results_seegull}) on Japanese text after unlearning, independent of the unlearning source language, indicating that the intervention meaningfully alters the model’s confidence and internal representations for this language.

\section{Translation Quality}
\label{sec:translation_quality}
We sampled 100 instances from the TOFU and SeeGULL datasets for each language and asked native speakers of those languages to evaluate the translations produced by Google Translate. The annotators confirmed that the translations were semantically accurate, with only minor stylistic adjustments suggested that did not alter the original meaning. It is also important to note that the sentences in both datasets are typically very short, which simplifies the translation process and reduces the likelihood of complex errors.

\begin{table*}[!ht]
\centering
\small
\resizebox{\textwidth}{!}{
\begin{tabular}{c|c|cccccccccccc}
\toprule
\textbf{Language} & \textbf{Metric} & \textbf{Finetuned} & \textbf{Retain} & \textbf{en} & \textbf{fr} & \textbf{fa} & \textbf{ar} & \textbf{hi} & \textbf{iw} & \textbf{id} & \textbf{ru} & \textbf{ja} & \textbf{ko} \\
\midrule\midrule

\multirow{4}{*}{en}
 & MU  & 0.58 & 0.59 & 0.52 & 0.53 & 0.56 & 0.55 & 0.55 & 0.56 & 0.54 & 0.55 & 0.56 & 0.56 \\
 & PR  & 0.98 & 0.98 & 0.74 & 0.91 & 0.92 & 0.91 & 0.94 & 0.92 & 0.88 & 0.89 & 0.92 & 0.93 \\
 & PF  & 0.98 & 0.09 & 0.00 & 0.32 & 0.67 & 0.68 & 0.78 & 0.63 & 0.34 & 0.45 & 0.69 & 0.29 \\
 & TRF  & 0.48 & 0.67 & 0.51 & 0.51 & 0.44 & 0.45 & 0.51 & 0.49 & 0.50 & 0.51 & 0.47 & 0.53 \\
\midrule
\multirow{4}{*}{fr}
 & MU  & 0.51 & 0.51 & 0.48 & 0.47 & 0.50 & 0.48 & 0.48 & 0.48 & 0.48 & 0.48 & 0.49 & 0.48 \\
 & PR  & 0.97 & 0.97 & 0.87 & 0.84 & 0.91 & 0.89 & 0.93 & 0.90 & 0.88 & 0.88 & 0.92 & 0.92 \\
 & PF  & 0.96 & 0.10 & 0.24 & 0.03 & 0.63 & 0.63 & 0.78 & 0.59 & 0.28 & 0.50 & 0.68 & 0.42 \\
 & TRF  & 0.48 & 0.69 & 0.53 & 0.61 & 0.53 & 0.53 & 0.53 & 0.52 & 0.55 & 0.56 & 0.53 & 0.56 \\
\midrule
\multirow{4}{*}{fa}
 & MU  & 0.43 & 0.44 & 0.43 & 0.42 & 0.42 & 0.43 & 0.42 & 0.42 & 0.42 & 0.42 & 0.42 & 0.42 \\
 & PR  & 0.94 & 0.94 & 0.87 & 0.87 & 0.70 & 0.83 & 0.86 & 0.83 & 0.83 & 0.83 & 0.86 & 0.86 \\
 & PF  & 0.91 & 0.10 & 0.65 & 0.64 & 0.00 & 0.53 & 0.63 & 0.59 & 0.60 & 0.60 & 0.68 & 0.67 \\
 & TRF  & 0.56 & 0.70 & 0.56 & 0.54 & 0.67 & 0.56 & 0.58 & 0.60 & 0.56 & 0.59 & 0.55 & 0.55 \\
\midrule
\multirow{4}{*}{ar}
 & MU  & 0.43 & 0.43 & 0.44 & 0.43 & 0.45 & 0.43 & 0.43 & 0.44 & 0.43 & 0.43 & 0.43 & 0.43 \\
 & PR  & 0.94 & 0.95 & 0.87 & 0.87 & 0.84 & 0.75 & 0.88 & 0.83 & 0.84 & 0.84 & 0.87 & 0.87 \\
 & PF  & 0.91 & 0.10 & 0.63 & 0.61 & 0.41 & 0.01 & 0.71 & 0.52 & 0.58 & 0.59 & 0.73 & 0.70 \\
 & TRF  & 0.51 & 0.64 & 0.48 & 0.48 & 0.49 & 0.52 & 0.53 & 0.52 & 0.52 & 0.54 & 0.49 & 0.46 \\
\midrule
\multirow{4}{*}{hi}
 & MU  & 0.39 & 0.40 & 0.40 & 0.40 & 0.41 & 0.41 & 0.41 & 0.41 & 0.41 & 0.40 & 0.41 & 0.41 \\
 & PR  & 0.97 & 0.97 & 0.92 & 0.93 & 0.91 & 0.91 & 0.83 & 0.90 & 0.91 & 0.91 & 0.91 & 0.91 \\
 & PF  & 0.98 & 0.31 & 0.86 & 0.86 & 0.75 & 0.88 & 0.04 & 0.84 & 0.82 & 0.80 & 0.75 & 0.78 \\
 & TRF  & 0.73 & 0.81 & 0.72 & 0.70 & 0.69 & 0.70 & 0.73 & 0.69 & 0.70 & 0.71 & 0.70 & 0.71 \\
\midrule
\multirow{4}{*}{iw}
 & MU  & 0.42 & 0.42 & 0.41 & 0.40 & 0.42 & 0.41 & 0.41 & 0.40 & 0.41 & 0.40 & 0.41 & 0.41 \\
 & PR  & 0.93 & 0.93 & 0.86 & 0.87 & 0.85 & 0.84 & 0.87 & 0.76 & 0.83 & 0.84 & 0.87 & 0.87 \\
 & PF  & 0.92 & 0.11 & 0.61 & 0.62 & 0.58 & 0.64 & 0.76 & 0.01 & 0.56 & 0.61 & 0.72 & 0.73 \\
 & TRF  & 0.57 & 0.73 & 0.57 & 0.57 & 0.57 & 0.55 & 0.58 & 0.66 & 0.59 & 0.58 & 0.59 & 0.57 \\
\midrule
\multirow{4}{*}{id}
 & MU  & 0.51 & 0.50 & 0.49 & 0.47 & 0.50 & 0.48 & 0.48 & 0.49 & 0.46 & 0.48 & 0.48 & 0.49 \\
 & PR  & 0.96 & 0.96 & 0.87 & 0.88 & 0.88 & 0.86 & 0.91 & 0.86 & 0.71 & 0.85 & 0.89 & 0.90 \\
 & PF  & 0.95 & 0.08 & 0.28 & 0.25 & 0.59 & 0.62 & 0.82 & 0.58 & 0.00 & 0.43 & 0.70 & 0.42 \\
 & TRF  & 0.48 & 0.66 & 0.54 & 0.52 & 0.45 & 0.47 & 0.48 & 0.47 & 0.53 & 0.53 & 0.46 & 0.53 \\
\midrule
\multirow{4}{*}{ru}
 & MU  & 0.44 & 0.45 & 0.43 & 0.42 & 0.43 & 0.43 & 0.43 & 0.42 & 0.43 & 0.41 & 0.42 & 0.42 \\
 & PR  & 0.93 & 0.93 & 0.84 & 0.86 & 0.85 & 0.83 & 0.87 & 0.84 & 0.83 & 0.72 & 0.86 & 0.87 \\
 & PF  & 0.90 & 0.08 & 0.45 & 0.45 & 0.52 & 0.64 & 0.69 & 0.55 & 0.50 & 0.01 & 0.66 & 0.58 \\
 & TRF  & 0.55 & 0.69 & 0.57 & 0.60 & 0.58 & 0.56 & 0.58 & 0.58 & 0.58 & 0.66 & 0.58 & 0.59 \\
\midrule
\multirow{4}{*}{ja}
 & MU  & 0.50 & 0.50 & 0.50 & 0.49 & 0.49 & 0.49 & 0.49 & 0.49 & 0.48 & 0.49 & 0.48 & 0.48 \\
 & PR  & 0.92 & 0.92 & 0.83 & 0.85 & 0.83 & 0.83 & 0.82 & 0.82 & 0.82 & 0.81 & 0.68 & 0.78 \\
 & PF  & 0.91 & 0.13 & 0.57 & 0.65 & 0.65 & 0.74 & 0.56 & 0.66 & 0.66 & 0.62 & 0.00 & 0.34 \\
 & TRF  & 0.62 & 0.74 & 0.64 & 0.61 & 0.60 & 0.60 & 0.63 & 0.62 & 0.61 & 0.64 & 0.56 & 0.62 \\
\midrule
\multirow{4}{*}{ko}
 & MU  & 0.47 & 0.49 & 0.47 & 0.46 & 0.46 & 0.46 & 0.46 & 0.46 & 0.46 & 0.46 & 0.46 & 0.45 \\
 & PR  & 0.92 & 0.92 & 0.84 & 0.85 & 0.82 & 0.82 & 0.82 & 0.81 & 0.82 & 0.81 & 0.77 & 0.70 \\
 & PF  & 0.93 & 0.10 & 0.30 & 0.41 & 0.67 & 0.75 & 0.63 & 0.69 & 0.50 & 0.63 & 0.29 & 0.00 \\
 & TRF  & 0.55 & 0.67 & 0.54 & 0.59 & 0.56 & 0.54 & 0.59 & 0.57 & 0.60 & 0.58 & 0.58 & 0.66 \\
\midrule
\bottomrule
\end{tabular}
}
\caption{Full results of unlearning experiments on the \textbf{TOFU} dataset using the \textbf{GradDiff} method across ten languages. Each row group corresponds to the evaluation language, while each column (after \textit{Finetuned} and \textit{Retain}) represents a model that has been unlearned on the respective language. Metrics include \textbf{Model Utility (MU)}, \textbf{Prob. Retain (PR)}, \textbf{Prob. Forget (PF)}, and \textbf{Truth Ratio Forget (TRF)}.}

\label{tab:grad_full}
\end{table*}

\begin{table*}[!ht]
\centering
\small
\resizebox{\textwidth}{!}{
\begin{tabular}{c|c|cccccccccccc}
\toprule
\textbf{Language} & \textbf{Metric} & \textbf{Finetuned} & \textbf{Retain} & \textbf{en} & \textbf{fr} & \textbf{fa} & \textbf{ar} & \textbf{hi} & \textbf{iw} & \textbf{id} & \textbf{ru} & \textbf{ja} & \textbf{ko} \\
\midrule\midrule

\multirow{4}{*}{en}
 & MU  & 0.58 & 0.59 & 0.52 & 0.54 & 0.56 & 0.54 & 0.53 & 0.54 & 0.54 & 0.55 & 0.55 & 0.53 \\
 & PR  & 0.98 & 0.98 & 0.76 & 0.89 & 0.89 & 0.92 & 0.93 & 0.90 & 0.81 & 0.85 & 0.90 & 0.91 \\
 & PF  & 0.98 & 0.09 & 0.00 & 0.12 & 0.56 & 0.48 & 0.71 & 0.27 & 0.06 & 0.19 & 0.43 & 0.12 \\
 & TRF  & 0.48 & 0.67 & 0.52 & 0.60 & 0.46 & 0.46 & 0.50 & 0.54 & 0.56 & 0.56 & 0.49 & 0.58 \\
\midrule
\multirow{4}{*}{fr}
 & MU  & 0.51 & 0.51 & 0.49 & 0.45 & 0.48 & 0.49 & 0.48 & 0.48 & 0.48 & 0.48 & 0.49 & 0.48 \\
 & PR  & 0.97 & 0.97 & 0.79 & 0.79 & 0.87 & 0.91 & 0.92 & 0.89 & 0.81 & 0.87 & 0.90 & 0.91 \\
 & PF  & 0.96 & 0.10 & 0.13 & 0.00 & 0.50 & 0.48 & 0.74 & 0.27 & 0.06 & 0.23 & 0.47 & 0.26 \\
 & TRF  & 0.48 & 0.69 & 0.58 & 0.61 & 0.53 & 0.48 & 0.55 & 0.53 & 0.58 & 0.57 & 0.55 & 0.58 \\
\midrule
\multirow{4}{*}{fa}
 & MU  & 0.43 & 0.44 & 0.43 & 0.42 & 0.42 & 0.43 & 0.41 & 0.42 & 0.43 & 0.42 & 0.42 & 0.42 \\
 & PR  & 0.94 & 0.94 & 0.87 & 0.85 & 0.61 & 0.85 & 0.85 & 0.81 & 0.79 & 0.82 & 0.84 & 0.85 \\
 & PF  & 0.91 & 0.10 & 0.59 & 0.52 & 0.00 & 0.41 & 0.54 & 0.39 & 0.35 & 0.31 & 0.47 & 0.57 \\
 & TRF  & 0.56 & 0.70 & 0.59 & 0.58 & 0.64 & 0.62 & 0.60 & 0.62 & 0.57 & 0.61 & 0.57 & 0.58 \\
\midrule
\multirow{4}{*}{ar}
 & MU  & 0.43 & 0.43 & 0.44 & 0.43 & 0.45 & 0.43 & 0.43 & 0.43 & 0.42 & 0.43 & 0.43 & 0.42 \\
 & PR  & 0.94 & 0.95 & 0.87 & 0.85 & 0.79 & 0.76 & 0.87 & 0.83 & 0.80 & 0.83 & 0.84 & 0.87 \\
 & PF  & 0.91 & 0.10 & 0.54 & 0.47 & 0.28 & 0.01 & 0.66 & 0.45 & 0.37 & 0.46 & 0.56 & 0.64 \\
 & TRF  & 0.51 & 0.64 & 0.51 & 0.48 & 0.53 & 0.55 & 0.54 & 0.49 & 0.51 & 0.53 & 0.50 & 0.46 \\
\midrule
\multirow{4}{*}{hi}
 & MU  & 0.39 & 0.40 & 0.40 & 0.41 & 0.40 & 0.41 & 0.40 & 0.41 & 0.42 & 0.41 & 0.41 & 0.40 \\
 & PR  & 0.97 & 0.97 & 0.92 & 0.92 & 0.88 & 0.92 & 0.74 & 0.90 & 0.90 & 0.90 & 0.88 & 0.91 \\
 & PF  & 0.98 & 0.31 & 0.79 & 0.83 & 0.63 & 0.81 & 0.03 & 0.65 & 0.66 & 0.48 & 0.54 & 0.71 \\
 & TRF  & 0.73 & 0.81 & 0.73 & 0.69 & 0.70 & 0.71 & 0.65 & 0.73 & 0.68 & 0.70 & 0.67 & 0.71 \\
\midrule
\multirow{4}{*}{iw}
 & MU  & 0.42 & 0.42 & 0.41 & 0.41 & 0.42 & 0.41 & 0.41 & 0.40 & 0.41 & 0.40 & 0.41 & 0.40 \\
 & PR  & 0.93 & 0.93 & 0.86 & 0.85 & 0.80 & 0.85 & 0.86 & 0.72 & 0.79 & 0.82 & 0.85 & 0.87 \\
 & PF  & 0.92 & 0.11 & 0.52 & 0.49 & 0.43 & 0.51 & 0.67 & 0.00 & 0.36 & 0.33 & 0.52 & 0.62 \\
 & TRF  & 0.57 & 0.73 & 0.58 & 0.59 & 0.60 & 0.56 & 0.63 & 0.65 & 0.60 & 0.62 & 0.60 & 0.58 \\
\midrule
\multirow{4}{*}{id}
 & MU  & 0.51 & 0.50 & 0.49 & 0.48 & 0.50 & 0.48 & 0.47 & 0.47 & 0.46 & 0.48 & 0.48 & 0.46 \\
 & PR  & 0.96 & 0.96 & 0.86 & 0.87 & 0.82 & 0.88 & 0.89 & 0.85 & 0.55 & 0.84 & 0.87 & 0.89 \\
 & PF  & 0.95 & 0.08 & 0.18 & 0.19 & 0.49 & 0.54 & 0.68 & 0.26 & 0.00 & 0.18 & 0.43 & 0.26 \\
 & TRF  & 0.48 & 0.66 & 0.62 & 0.56 & 0.47 & 0.49 & 0.50 & 0.50 & 0.51 & 0.53 & 0.49 & 0.53 \\
\midrule
\multirow{4}{*}{ru}
 & MU  & 0.44 & 0.45 & 0.44 & 0.43 & 0.44 & 0.43 & 0.41 & 0.41 & 0.44 & 0.40 & 0.42 & 0.41 \\
 & PR  & 0.93 & 0.93 & 0.84 & 0.84 & 0.80 & 0.86 & 0.86 & 0.82 & 0.77 & 0.53 & 0.83 & 0.87 \\
 & PF  & 0.90 & 0.08 & 0.35 & 0.32 & 0.35 & 0.50 & 0.60 & 0.24 & 0.20 & 0.00 & 0.43 & 0.44 \\
 & TRF  & 0.55 & 0.69 & 0.60 & 0.59 & 0.60 & 0.56 & 0.57 & 0.60 & 0.60 & 0.61 & 0.55 & 0.59 \\
\midrule
\multirow{4}{*}{ja}
 & MU  & 0.50 & 0.50 & 0.50 & 0.50 & 0.49 & 0.49 & 0.49 & 0.49 & 0.49 & 0.49 & 0.47 & 0.48 \\
 & PR  & 0.92 & 0.92 & 0.84 & 0.83 & 0.79 & 0.85 & 0.80 & 0.81 & 0.81 & 0.79 & 0.57 & 0.76 \\
 & PF  & 0.91 & 0.13 & 0.50 & 0.58 & 0.49 & 0.62 & 0.46 & 0.47 & 0.42 & 0.37 & 0.00 & 0.27 \\
 & TRF  & 0.62 & 0.74 & 0.59 & 0.62 & 0.62 & 0.61 & 0.62 & 0.57 & 0.59 & 0.64 & 0.48 & 0.60 \\
\midrule
\multirow{4}{*}{ko}
 & MU  & 0.47 & 0.49 & 0.47 & 0.47 & 0.46 & 0.47 & 0.45 & 0.45 & 0.47 & 0.46 & 0.46 & 0.43 \\
 & PR  & 0.92 & 0.92 & 0.84 & 0.83 & 0.78 & 0.85 & 0.80 & 0.81 & 0.79 & 0.79 & 0.72 & 0.65 \\
 & PF  & 0.93 & 0.10 & 0.22 & 0.27 & 0.55 & 0.62 & 0.51 & 0.37 & 0.20 & 0.34 & 0.18 & 0.00 \\
 & TRF  & 0.55 & 0.67 & 0.59 & 0.59 & 0.59 & 0.57 & 0.58 & 0.60 & 0.60 & 0.59 & 0.61 & 0.62 \\
\midrule
\bottomrule
\end{tabular}
}
\caption{Full results of unlearning experiments on the \textbf{TOFU} dataset using the \textbf{GradDiff-KL} method across ten languages. Each row group corresponds to the evaluation language, while each column (after \textit{Finetuned} and \textit{Retain}) represents a model that has been unlearned on the respective language. Metrics include \textbf{Model Utility (MU)}, \textbf{Prob. Retain (PR)}, \textbf{Prob. Forget (PF)}, and \textbf{Truth Ratio Forget (TRF)}.}
\label{tab:kl_full}
\end{table*}

\begin{table*}[!ht]
\centering
\small
\resizebox{\textwidth}{!}{
\begin{tabular}{c|c|cccccccccccc}
\toprule
\textbf{Language} & \textbf{Metric} & \textbf{Finetuned} & \textbf{Retain} & \textbf{en} & \textbf{fr} & \textbf{fa} & \textbf{ar} & \textbf{hi} & \textbf{iw} & \textbf{id} & \textbf{ru} & \textbf{ja} & \textbf{ko} \\
\midrule\midrule

\multirow{4}{*}{en}
 & MU  & 0.58 & 0.59 & 0.61 & 0.59 & 0.58 & 0.58 & 0.58 & 0.58 & 0.59 & 0.58 & 0.58 & 0.59 \\
 & PR  & 0.98 & 0.98 & 0.96 & 0.97 & 0.97 & 0.97 & 0.97 & 0.97 & 0.97 & 0.97 & 0.97 & 0.97 \\
 & PF  & 0.98 & 0.09 & 0.25 & 0.71 & 0.92 & 0.91 & 0.94 & 0.88 & 0.80 & 0.83 & 0.93 & 0.83 \\
 & TRF  & 0.48 & 0.67 & 0.53 & 0.50 & 0.49 & 0.47 & 0.49 & 0.49 & 0.50 & 0.50 & 0.48 & 0.51 \\
\midrule
\multirow{4}{*}{fr}
 & MU  & 0.51 & 0.51 & 0.52 & 0.51 & 0.51 & 0.51 & 0.51 & 0.51 & 0.51 & 0.51 & 0.51 & 0.51 \\
 & PR  & 0.97 & 0.97 & 0.96 & 0.95 & 0.96 & 0.96 & 0.97 & 0.96 & 0.96 & 0.96 & 0.97 & 0.97 \\
 & PF  & 0.96 & 0.10 & 0.68 & 0.33 & 0.87 & 0.88 & 0.93 & 0.85 & 0.76 & 0.84 & 0.91 & 0.84 \\
 & TRF  & 0.48 & 0.69 & 0.51 & 0.56 & 0.51 & 0.49 & 0.51 & 0.50 & 0.51 & 0.52 & 0.50 & 0.52 \\
\midrule
\multirow{4}{*}{fa}
 & MU  & 0.43 & 0.44 & 0.44 & 0.43 & 0.43 & 0.43 & 0.43 & 0.43 & 0.43 & 0.43 & 0.43 & 0.43 \\
 & PR  & 0.94 & 0.94 & 0.93 & 0.93 & 0.91 & 0.93 & 0.93 & 0.93 & 0.93 & 0.93 & 0.93 & 0.93 \\
 & PF  & 0.91 & 0.10 & 0.86 & 0.83 & 0.23 & 0.82 & 0.83 & 0.83 & 0.85 & 0.83 & 0.87 & 0.87 \\
 & TRF  & 0.56 & 0.70 & 0.58 & 0.57 & 0.67 & 0.58 & 0.59 & 0.57 & 0.57 & 0.58 & 0.56 & 0.56 \\
\midrule
\multirow{4}{*}{ar}
 & MU  & 0.43 & 0.43 & 0.44 & 0.43 & 0.43 & 0.44 & 0.43 & 0.43 & 0.43 & 0.43 & 0.43 & 0.43 \\
 & PR  & 0.94 & 0.95 & 0.93 & 0.93 & 0.93 & 0.92 & 0.94 & 0.93 & 0.93 & 0.93 & 0.94 & 0.94 \\
 & PF  & 0.91 & 0.10 & 0.83 & 0.82 & 0.79 & 0.26 & 0.86 & 0.83 & 0.85 & 0.84 & 0.88 & 0.88 \\
 & TRF  & 0.51 & 0.64 & 0.54 & 0.51 & 0.52 & 0.54 & 0.53 & 0.52 & 0.53 & 0.52 & 0.52 & 0.51 \\
\midrule
\multirow{4}{*}{hi}
 & MU  & 0.39 & 0.40 & 0.39 & 0.39 & 0.39 & 0.40 & 0.39 & 0.39 & 0.39 & 0.39 & 0.39 & 0.39 \\
 & PR  & 0.97 & 0.97 & 0.97 & 0.97 & 0.97 & 0.97 & 0.96 & 0.97 & 0.97 & 0.97 & 0.97 & 0.97 \\
 & PF  & 0.98 & 0.31 & 0.96 & 0.96 & 0.94 & 0.96 & 0.56 & 0.96 & 0.96 & 0.95 & 0.94 & 0.95 \\
 & TRF  & 0.73 & 0.81 & 0.74 & 0.73 & 0.73 & 0.72 & 0.77 & 0.73 & 0.74 & 0.73 & 0.73 & 0.75 \\
\midrule
\multirow{4}{*}{iw}
 & MU  & 0.42 & 0.42 & 0.42 & 0.42 & 0.42 & 0.42 & 0.42 & 0.42 & 0.42 & 0.42 & 0.42 & 0.42 \\
 & PR  & 0.93 & 0.93 & 0.92 & 0.92 & 0.92 & 0.92 & 0.92 & 0.91 & 0.92 & 0.92 & 0.92 & 0.93 \\
 & PF  & 0.92 & 0.11 & 0.81 & 0.81 & 0.85 & 0.86 & 0.89 & 0.26 & 0.85 & 0.85 & 0.89 & 0.89 \\
 & TRF  & 0.57 & 0.73 & 0.59 & 0.57 & 0.58 & 0.57 & 0.58 & 0.61 & 0.58 & 0.58 & 0.58 & 0.57 \\
\midrule
\multirow{4}{*}{id}
 & MU  & 0.51 & 0.50 & 0.52 & 0.52 & 0.51 & 0.52 & 0.51 & 0.51 & 0.52 & 0.51 & 0.50 & 0.51 \\
 & PR  & 0.96 & 0.96 & 0.94 & 0.95 & 0.95 & 0.95 & 0.95 & 0.95 & 0.94 & 0.95 & 0.95 & 0.95 \\
 & PF  & 0.95 & 0.08 & 0.75 & 0.72 & 0.85 & 0.88 & 0.92 & 0.84 & 0.29 & 0.85 & 0.91 & 0.84 \\
 & TRF  & 0.48 & 0.66 & 0.51 & 0.52 & 0.49 & 0.49 & 0.50 & 0.48 & 0.53 & 0.51 & 0.48 & 0.51 \\
\midrule
\multirow{4}{*}{ru}
 & MU  & 0.44 & 0.45 & 0.46 & 0.45 & 0.44 & 0.45 & 0.44 & 0.44 & 0.45 & 0.44 & 0.44 & 0.45 \\
 & PR  & 0.93 & 0.93 & 0.92 & 0.92 & 0.92 & 0.92 & 0.93 & 0.92 & 0.92 & 0.92 & 0.93 & 0.93 \\
 & PF  & 0.90 & 0.08 & 0.76 & 0.74 & 0.82 & 0.85 & 0.85 & 0.81 & 0.81 & 0.24 & 0.86 & 0.83 \\
 & TRF  & 0.55 & 0.69 & 0.56 & 0.56 & 0.56 & 0.55 & 0.56 & 0.55 & 0.56 & 0.59 & 0.56 & 0.55 \\
\midrule
\multirow{4}{*}{ja}
 & MU  & 0.50 & 0.50 & 0.51 & 0.50 & 0.50 & 0.50 & 0.50 & 0.50 & 0.50 & 0.50 & 0.50 & 0.50 \\
 & PR  & 0.92 & 0.92 & 0.91 & 0.91 & 0.91 & 0.92 & 0.91 & 0.91 & 0.91 & 0.91 & 0.90 & 0.91 \\
 & PF  & 0.91 & 0.13 & 0.81 & 0.83 & 0.86 & 0.88 & 0.82 & 0.88 & 0.86 & 0.86 & 0.25 & 0.76 \\
 & TRF  & 0.62 & 0.74 & 0.63 & 0.62 & 0.62 & 0.59 & 0.62 & 0.61 & 0.62 & 0.63 & 0.65 & 0.63 \\
\midrule
\multirow{4}{*}{ko}
 & MU  & 0.47 & 0.49 & 0.48 & 0.47 & 0.47 & 0.48 & 0.47 & 0.47 & 0.48 & 0.47 & 0.47 & 0.48 \\
 & PR  & 0.92 & 0.92 & 0.91 & 0.91 & 0.91 & 0.91 & 0.91 & 0.91 & 0.91 & 0.91 & 0.91 & 0.90 \\
 & PF  & 0.93 & 0.10 & 0.68 & 0.73 & 0.89 & 0.90 & 0.84 & 0.88 & 0.83 & 0.86 & 0.79 & 0.26 \\
 & TRF  & 0.55 & 0.67 & 0.56 & 0.56 & 0.55 & 0.54 & 0.55 & 0.55 & 0.56 & 0.55 & 0.54 & 0.59 \\
\midrule
\bottomrule
\end{tabular}
}
\caption{Full results of unlearning experiments on the \textbf{TOFU} dataset using the \textbf{NPO} method across ten languages. Each row group corresponds to the evaluation language, while each column (after \textit{Finetuned} and \textit{Retain}) represents a model that has been unlearned on the respective language. Metrics include \textbf{Model Utility (MU)}, \textbf{Prob. Retain (PR)}, \textbf{Prob. Forget (PF)}, and \textbf{Truth Ratio Forget (TRF)}.}
\label{tab:npo_full}
\end{table*}

\begin{figure*}[t]
    \centering
    \begin{subfigure}[b]{0.49\linewidth}
        \centering
        \includegraphics[width=\linewidth]{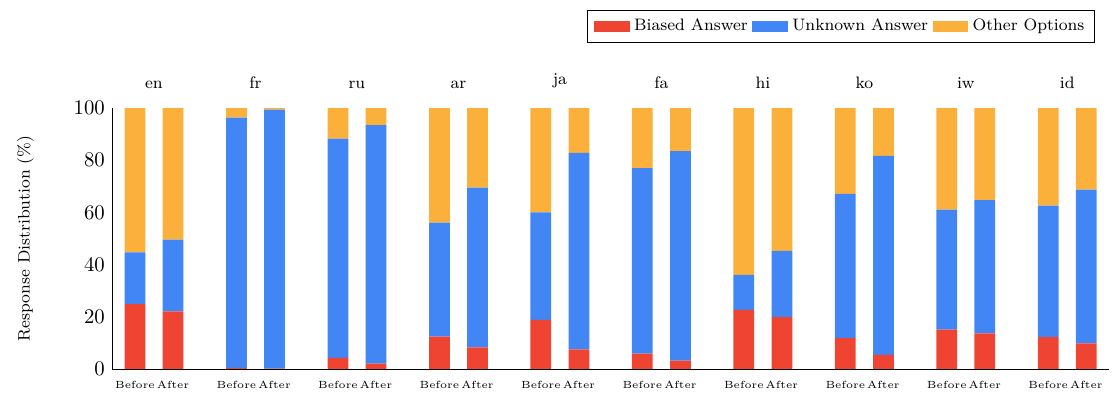}
        \caption{GradDiff-KL (unlearned on \texttt{fr})}
        \label{fig:grad_seegull_fr}
    \end{subfigure}
    \hfill
    \begin{subfigure}[b]{0.49\linewidth}
        \centering
        \includegraphics[width=\linewidth]{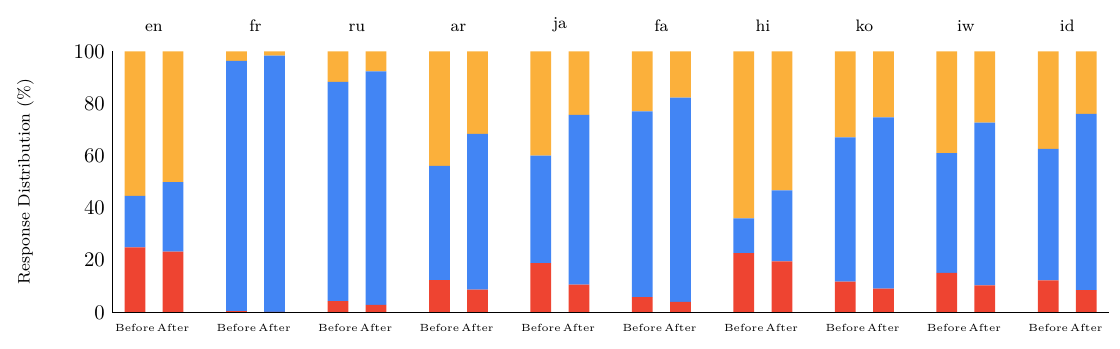}
        \caption{NPO (unlearned on \texttt{fr})}
        \label{fig:npo_seegull_fr}
    \end{subfigure}
    \par\medskip

    \begin{subfigure}[b]{0.49\linewidth}
        \centering
        \includegraphics[width=\linewidth]{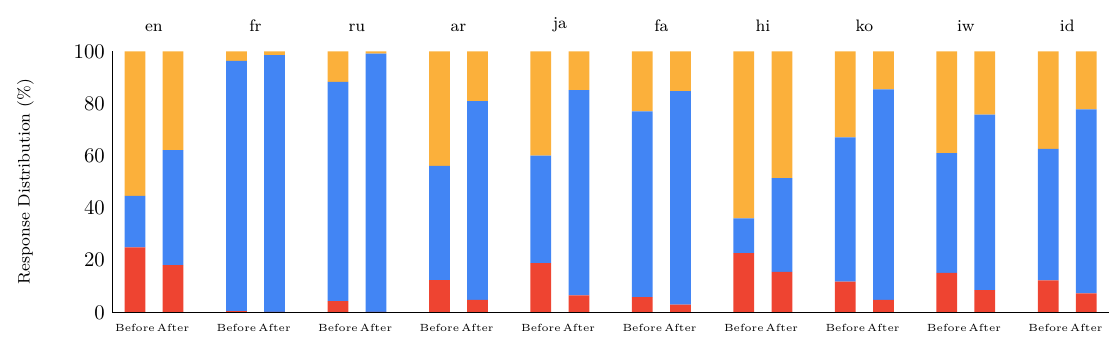}
        \caption{GradDiff-KL (unlearned on \texttt{ru})}
        \label{fig:grad_seegull_ru}
    \end{subfigure}
    \hfill
    \begin{subfigure}[b]{0.49\linewidth}
        \centering
        \includegraphics[width=\linewidth]{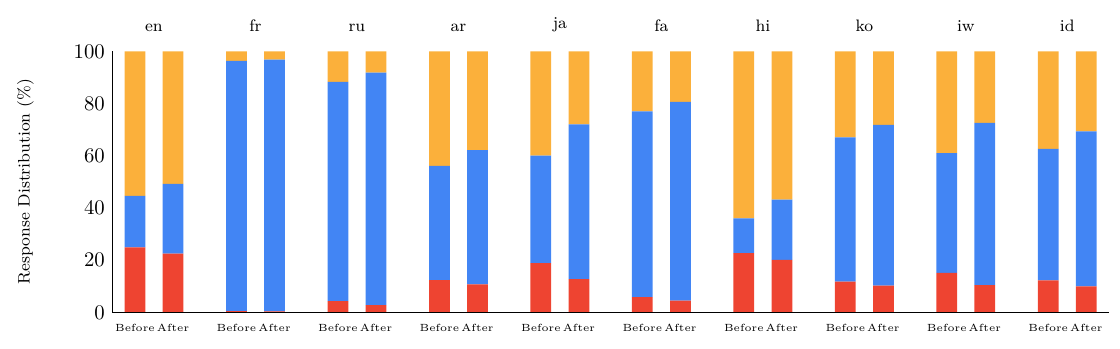}
        \caption{NPO (unlearned on \texttt{ru})}
        \label{fig:npo_seegull_ru}
    \end{subfigure}
    \par\medskip

    \begin{subfigure}[b]{0.49\linewidth}
        \centering
        \includegraphics[width=\linewidth]{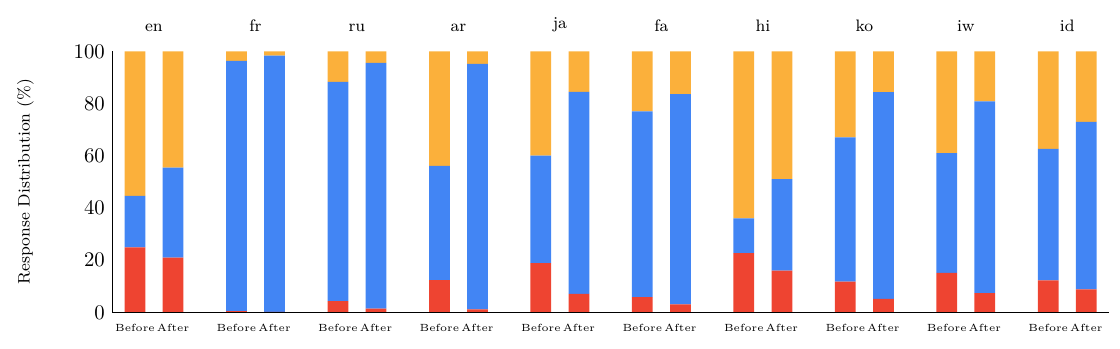}
        \caption{GradDiff-KL (unlearned on \texttt{ar})}
        \label{fig:grad_seegull_ar}
    \end{subfigure}
    \hfill
    \begin{subfigure}[b]{0.49\linewidth}
        \centering
        \includegraphics[width=\linewidth]{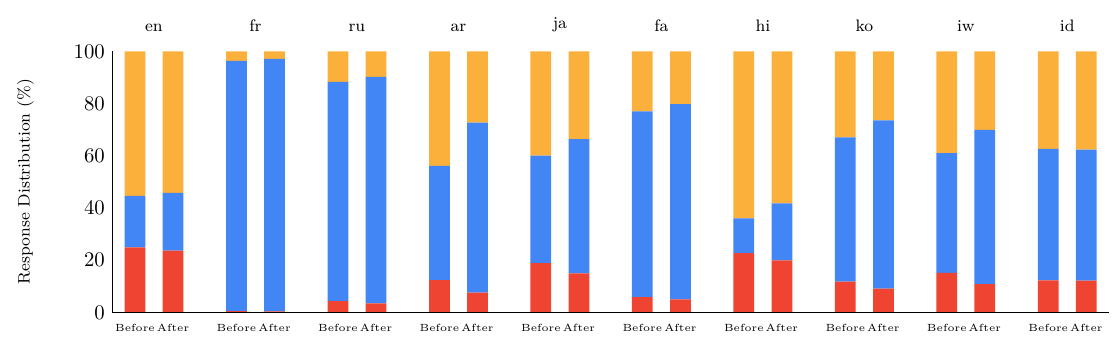}
        \caption{NPO (unlearned on \texttt{ar})}
        \label{fig:npo_seegull_ar}
    \end{subfigure}
    \par\medskip

    \begin{subfigure}[b]{0.49\linewidth}
        \centering
        \includegraphics[width=\linewidth]{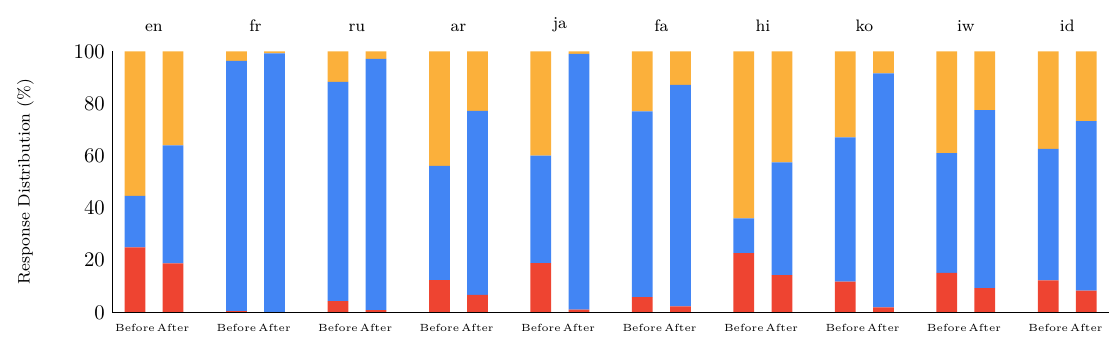}
        \caption{GradDiff-KL (unlearned on \texttt{ja})}
        \label{fig:grad_seegull_ja}
    \end{subfigure}
    \hfill
    \begin{subfigure}[b]{0.49\linewidth}
        \centering
        \includegraphics[width=\linewidth]{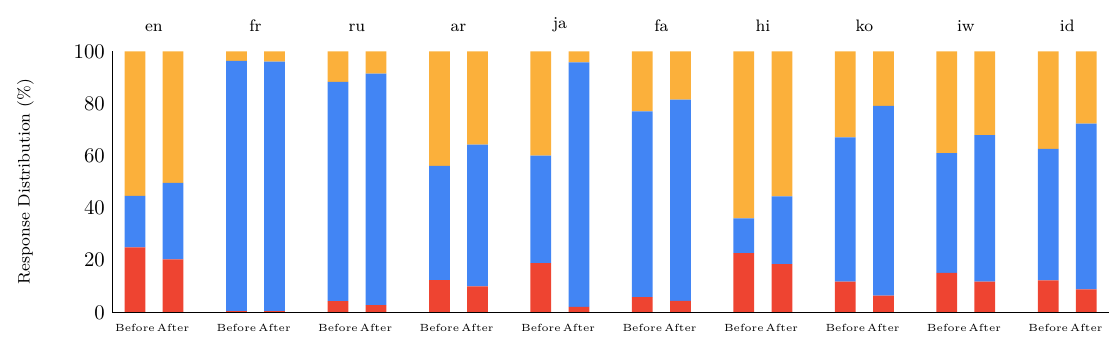}
        \caption{NPO (unlearned on \texttt{ja})}
        \label{fig:npo_seegull_ja}
    \end{subfigure}
    \par\medskip

    \begin{subfigure}[b]{0.49\linewidth}
        \centering
        \includegraphics[width=\linewidth]{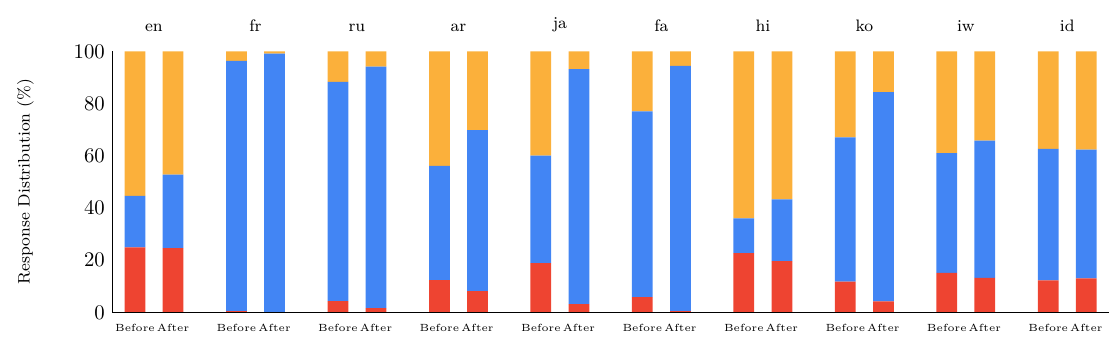}
        \caption{GradDiff-KL (unlearned on \texttt{fa})}
        \label{fig:grad_seegull_fa}
    \end{subfigure}
    \hfill
    \begin{subfigure}[b]{0.49\linewidth}
        \centering
        \includegraphics[width=\linewidth]{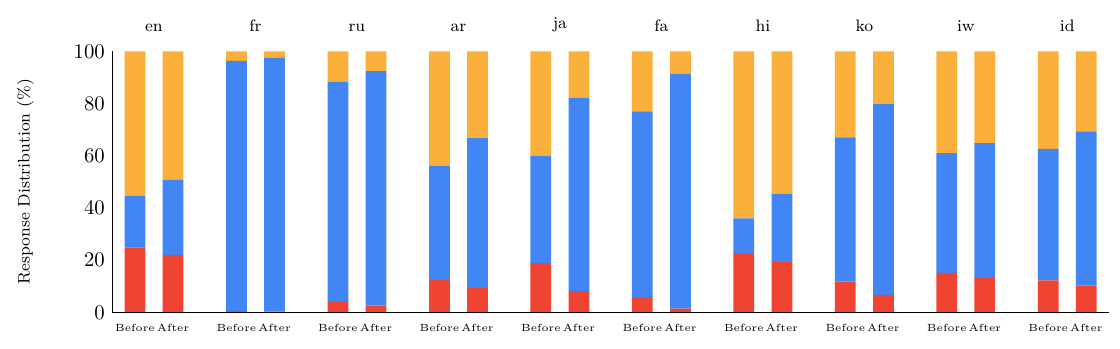}
        \caption{NPO (unlearned on \texttt{fa})}
        \label{fig:npo_seegull_fa}
    \end{subfigure}
    \par\medskip

\end{figure*}

\begin{figure*}[ht]\ContinuedFloat
    \begin{subfigure}[b]{0.49\linewidth}
        \centering
        \includegraphics[width=\linewidth]{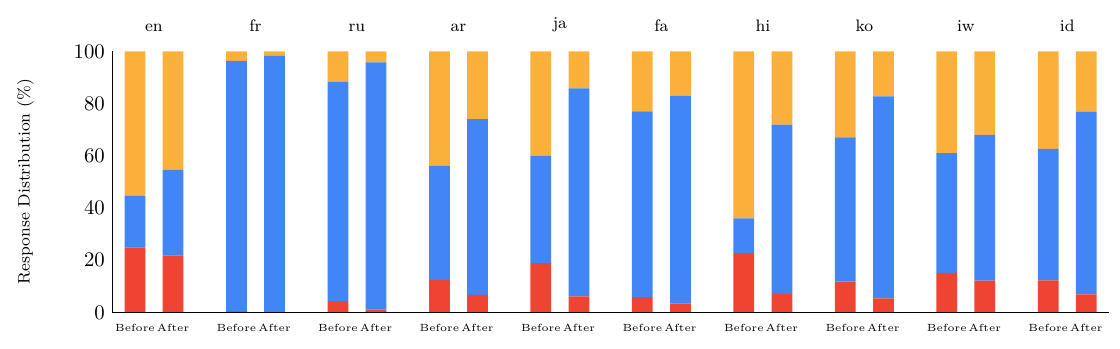}
        \caption{GradDiff-KL (unlearned on \texttt{hi})}
        \label{fig:grad_seegull_hi}
    \end{subfigure}
    \hfill
    \begin{subfigure}[b]{0.49\linewidth}
        \centering
        \includegraphics[width=\linewidth]{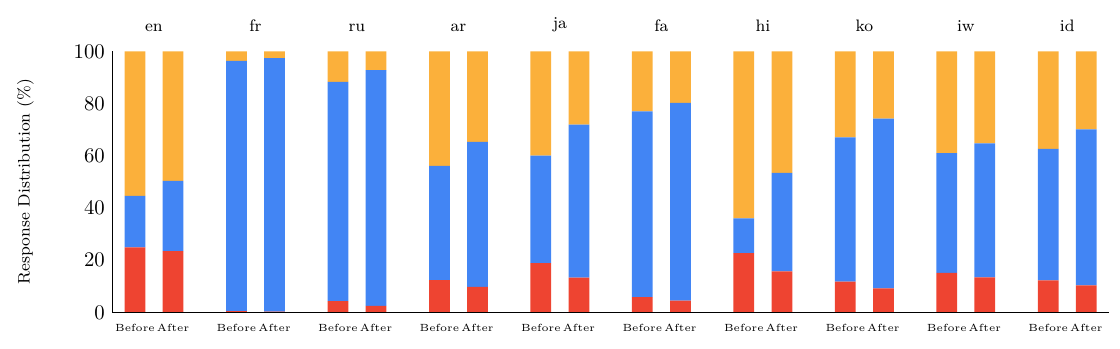}
        \caption{NPO (unlearned on \texttt{hi})}
        \label{fig:npo_seegull_hi}
    \end{subfigure}
    \par\medskip

    \begin{subfigure}[b]{0.49\linewidth}
        \centering
        \includegraphics[width=\linewidth]{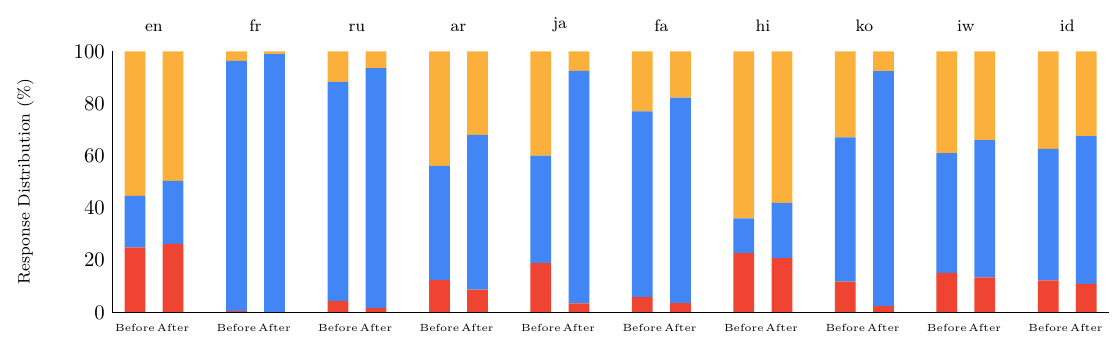}
        \caption{GradDiff-KL (unlearned on \texttt{ko})}
        \label{fig:grad_seegull_ko}
    \end{subfigure}
    \hfill
    \begin{subfigure}[b]{0.49\linewidth}
        \centering
        \includegraphics[width=\linewidth]{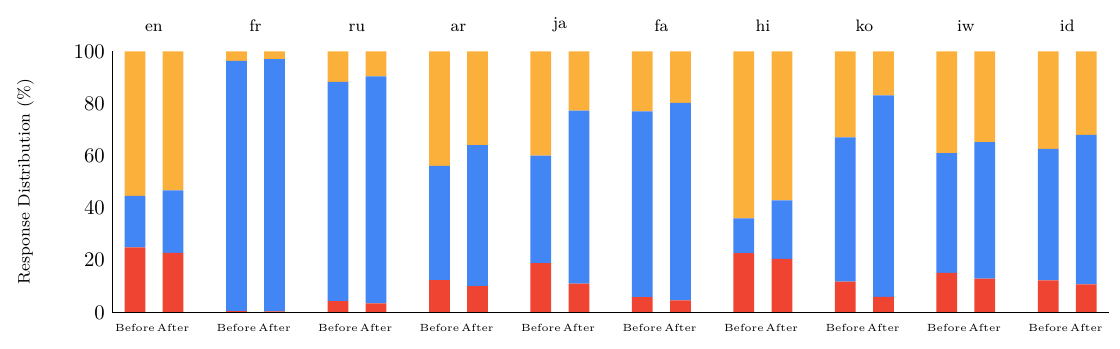}
        \caption{NPO (unlearned on \texttt{ko})}
        \label{fig:npo_seegull_ko}
    \end{subfigure}
    \par\medskip

    \begin{subfigure}[b]{0.49\linewidth}
        \centering
        \includegraphics[width=\linewidth]{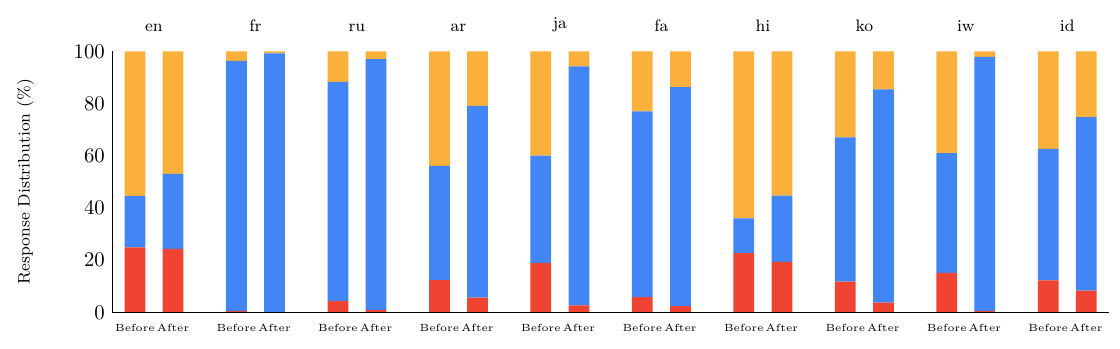}
        \caption{GradDiff-KL (unlearned on \texttt{iw})}
        \label{fig:grad_seegull_iw}
    \end{subfigure}
    \hfill
    \begin{subfigure}[b]{0.49\linewidth}
        \centering
        \includegraphics[width=\linewidth]{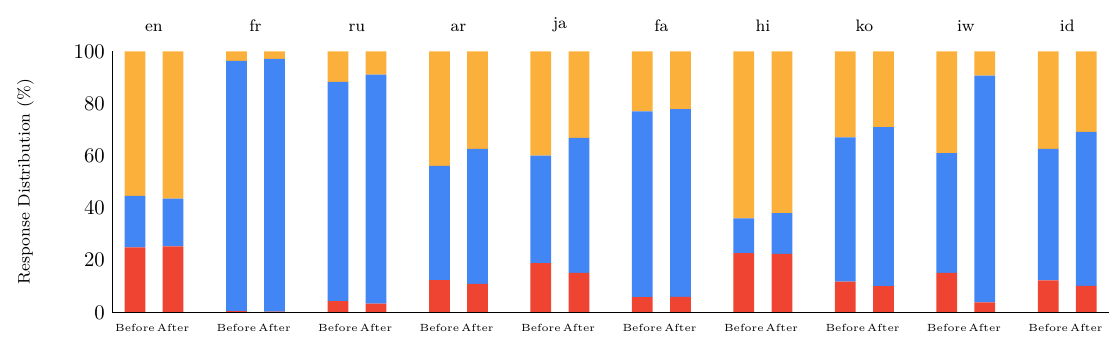}
        \caption{NPO (unlearned on \texttt{iw})}
        \label{fig:npo_seegull_iw}
    \end{subfigure}
    \par\medskip

    \begin{subfigure}[b]{0.49\linewidth}
        \centering
        \includegraphics[width=\linewidth]{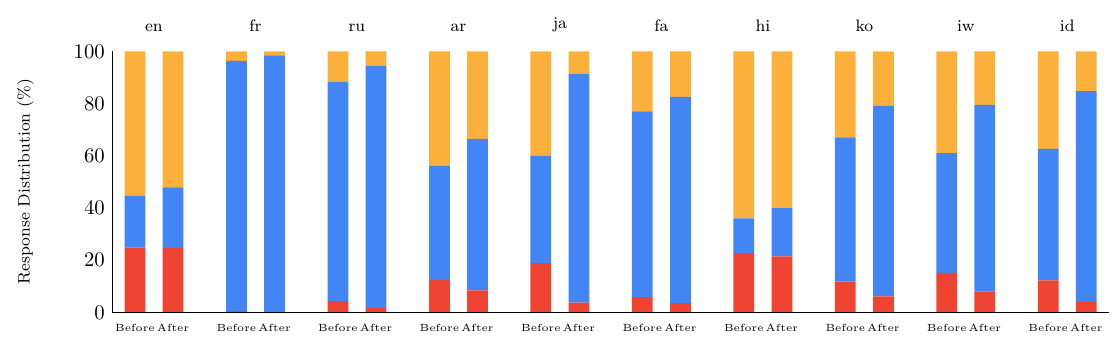}
        \caption{GradDiff-KL (unlearned on \texttt{id})}
        \label{fig:grad_seegull_id}
    \end{subfigure}
    \hfill
    \begin{subfigure}[b]{0.49\linewidth}
        \centering
        \includegraphics[width=\linewidth]{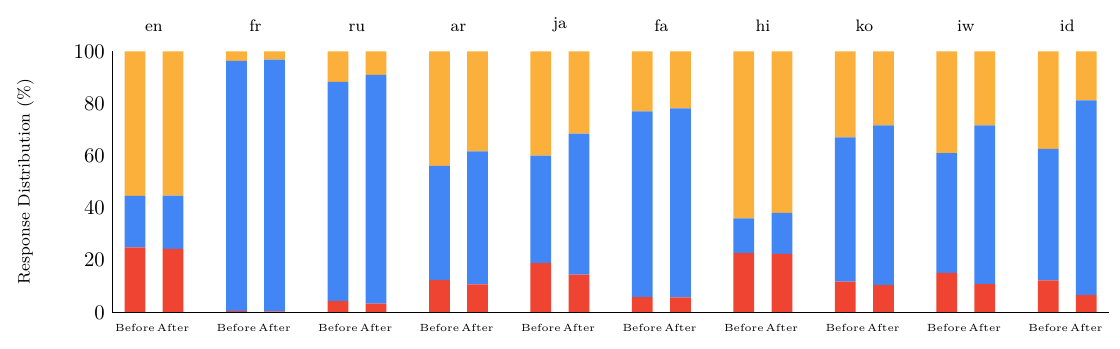}
        \caption{NPO (unlearned on \texttt{id})}
        \label{fig:npo_seegull_id}
    \end{subfigure}
    \vspace{-0.08in}
    \caption{Results on the SeeGULL QA dataset across nine languages (excluding English) before and after unlearning. Each row shows GradDiff-KL (left) and NPO (right) for the specified unlearning language.}
\end{figure*}

\end{document}